%% file: medima-template.tex
\documentclass[times,twocolumn,final]{elsarticle}

\usepackage{medima}
\usepackage{framed,multirow}

\usepackage{amssymb}
\usepackage{latexsym}

\usepackage{url}
\usepackage{xcolor}

\usepackage{hyperref}
\hypersetup{draft} 

\definecolor{newcolor}{rgb}{.8,.349,.1}

\renewcommand{\cite}[1]{\citep{#1}}

\usepackage{amsmath}            
\usepackage{epstopdf}           
\usepackage{flushend}           

\usepackage{lineno,hyperref}

\usepackage{paralist}
\usepackage{lipsum}
\usepackage{amsmath,amsfonts,amssymb}
\usepackage{xcolor}
\usepackage[font=footnotesize,labelfont=bf]{caption}
\usepackage[font=footnotesize,labelfont=bf]{subcaption}

\usepackage{longtable}
\usepackage{array}

\newcommand{\upper}[1]{$^{\text{#1}}$}
\newcommand{\pA}{Part~\textbf{A}}
\newcommand{\pB}{Part~\textbf{B}}
\newcommand{\pAB}{Part~\textbf{A}~and~\textbf{B}}
\newcommand{\etal}[1]{#1~\textit{et al.}}
\newcommand{\splito}[2]{\parbox[c][5.5ex][c]{5.5ex}{\centering #1\\#2}}

\usepackage{xcolor,soul}

\definecolor{green1}{RGB}{204,235,197}

\definecolor{orange1}{RGB}{254,217,166}
\definecolor{blue1}{RGB}{179,205,227}
\definecolor{pink1}{RGB}{251,180,174}
\definecolor{purple1}{RGB}{222,203,228}

\newcommand{\hlo}[1]{{\sethlcolor{orange1}\hl{#1}}}

\soulregister{\code}{1}
\setstcolor{red}

\usepackage{eurosym}
\usepackage{array}
\usepackage{amssymb}
\usepackage{pifont}
\newcommand{\cmark}{\ding{51}}%
\newcommand{\xmark}{\ding{55}}%
\usepackage{longtable}
\usepackage[figuresright]{rotating}
\usepackage{arydshln}

\journal{Medical Image Analysis}

\begin{document}

\verso{Guilherme Aresta \textit{et~al.}}

\begin{frontmatter}


\title{BACH: grand challenge on breast cancer histology images}
\tnotetext[mytitlenote]{$\copyright$ 2018. Licensed under the Creative Commons CC-BY-NC-ND 4.0 license \url{http://creativecommons.org/licenses/by-nc-nd/4.0/.}}

\address[inesc]{INESC TEC - Institute for Systems and Computer Engineering, Technology and Science, 4200-465 Porto, Portugal}
\address[feup]{Faculty of Engineering of University of Porto, 4200-465 Porto, Portugal}

\author[inesc,feup]{Guilherme Aresta\corref{equalcontribution}}
\cortext[equalcontribution]{Equal contribution; corresponding authors}

\ead{guilherme.m.aresta@inesctec.pt}

\author[inesc,feup]{Teresa Ara\'{u}jo\corref{equalcontribution}}
\ead{tfaraujo@inesctec.pt}

\author[AFkwok]{Scotty Kwok}
\address[AFkwok]{Seek AI Limited, Hong Kong, China}

\author[AFchenn1]{Sai Saketh Chennamsetty}
\author[AFchenn2]{Mohammed Safwan}
\author[AFchenn3]{Varghese Alex}
\address[AFchenn1]{Bangalore, India}
\address[AFchenn2]{Gurgaon, India}
\address[AFchenn3]{Chennai, India}

\author[AFmarami]{Bahram Marami}
\author[AFmarami]{Marcel Prastawa}
\author[AFmarami]{Monica Chan}
\author[AFmarami]{Michael Donovan}
\author[AFmarami]{Gerardo Fernandez}
\author[AFmarami]{Jack Zeineh}
\address[AFmarami]{The Center for Computational and Systems Pathology, Dpt. of Pathology, Icahn School of Medicine at Mount Sinai and The Mount Sinai Hospital, New York, USA}

\author[AFkohl1]{Matthias Kohl}
\author[AFkohl2]{Christoph Walz}
\author[AFkohl1]{Florian Ludwig}
\author[AFkohl1]{Stefan Braunewell}
\author[AFkohl1]{Maximilian Baust}
\address[AFkohl1]{Konica Minolta Laboratory Europe, Munich, Germany}
\address[AFkohl2]{Institute of Pathology, Faculty of Medicine, LMU Munich, Munich, Germany}

\author[AFvu]{Quoc Dang Vu}
\author[AFvu]{Minh Nguyen Nhat To}
\author[AFvu]{Eal Kim}
\author[AFvu]{Jin Tae Kwak}
\address[AFvu]{Department of Computer Science and Engineering, Sejong University, Seoul 05006, Korea}

\author[AFgalal]{Sameh Galal}
\author[AFgalal]{Veronica Sanchez-Freire}
\address[AFgalal]{Chicago, IL, USA}
\author[AFBrancati1]{Nadia Brancati}
\author[AFBrancati1]{Maria Frucci}
\author[AFBrancati1,AFBrancati2]{Daniel Riccio}
\address[AFBrancati1]{Institute for High Performance Computing and Networking, National Research Council of Italy (ICAR-CNR), Naples, Italy}
\address[AFBrancati2]{University of Naples ``Federico II", Naples, Italy}

\author[AFWang1]{Yaqi Wang}
\author[AFWang1,AFWang2]{Lingling Sun}
\author[AFWang1]{Kaiqiang Ma}
\author[AFWang1]{Jiannan Fang}
\address[AFWang1]{Key Laboratory of RF Circuits and Systems, Ministry of Education,
Hangzhou Dianzi University, Hangzhou 310018, China}
\address[AFWang2]{Zhejiang Provincial Laboratory of Integrated Circuits Design, Hangzhou Dianzi University, Hangzhou 310018, China}

\author[AFKone]{Ismael Kone}
\author[AFKone]{Lahsen Boulmane}
\address[AFKone]{2MIA Research Group, LEM2A Lab, Facult\' {e} des Sciences, Universit\'{e} Moulay Ismail, Meknes, Morocco}

\author[inesc,feup]{Aur\'{e}lio Campilho}

\author[ipatimup,fmup,i3s]{Catarina Eloy}
\address[ipatimup]{Laborat\'{o}rio de Anatomia Patol\'{o}gica, Ipatimup Diagn\'{o}sticos, Rua J\'{u}ilio Amaral de Carvalho, 45, 4200-135 Porto, Portugal}
\address[fmup]{Faculdade de Medicina, Universidade do Porto, Alameda Prof Hern\^{a}ni Monteiro, 4200-319 Porto, Portugal}

\author[ipatimup,fmup,i3s]{Ant\'{o}nio Pol\'{o}nia}
\ead{apolonia@ipatimup.pt}

\author[i3s,ineb]{Paulo Aguiar}
\ead{pauloaguiar@ineb.up.pt}
\address[i3s]{Instituto de Investiga\c{c}\~{a}o e Inova\c{c}\~{a}o em Sa\'{u}de (i3S), Universidade do Porto, Rua Alfredo Allen, 208, 4200-135 Porto, Portugal}
\address[ineb]{Instituto de Engenharia Biom\'{e}dica (INEB), Universidade do Porto, Rua Alfredo Allen, 208, 4200-135 Porto, Portugal}




\begin{abstract}
Breast cancer is the most common invasive cancer in women, affecting more than 10\% of women worldwide. Microscopic analysis of a biopsy remains one of the most important methods to diagnose the type of breast cancer. This requires specialized analysis by pathologists, in a task that i) is highly time- and cost-consuming and ii) often leads to nonconsensual results. The relevance and potential of automatic classification algorithms using hematoxylin-eosin stained histopathological images has already been demonstrated, but the reported results are still sub-optimal for clinical use. With the goal of advancing the state-of-the-art in automatic classification, the Grand Challenge on BreAst Cancer Histology images (BACH) was organized in conjunction with the 15th International Conference on Image Analysis and Recognition (ICIAR 2018).
BACH aimed at the classification and localization of clinically relevant histopathological classes in microscopy and whole-slide images from a large annotated dataset, specifically compiled and made publicly available for the challenge. Following a positive response from the scientific community, a total of 64 submissions, out of 677 registrations, effectively entered the competition. The submitted algorithms improved the state-of-the-art in automatic classification of breast cancer with microscopy images to an accuracy of 87\%.  Convolutional neuronal networks were the most successful methodology in the BACH challenge. Detailed analysis of the collective results allowed the identification of remaining challenges in the field and recommendations for future developments. The BACH dataset remains publicly available as to promote further improvements to the field of automatic classification in digital pathology.
\end{abstract}


\end{frontmatter}


\input{introduction}

\input{materials}

\input{methods}

\input{results}

\input{discussion}

\input{conclusion}

\section{Acknowledgments}

Guilherme Aresta is funded by the FCT grant contract \\ SFRH/BD/120435/2016. Teresa Ara\'{u}jo is funded by the FCT grant contract SFRH/BD/122365/2016. Aur\'{e}lio Campilho is with the project "NanoSTIMA: Macro-to-Nano Human Sensing: Towards Integrated Multimodal Health Monitoring and Analytics/NORTE-01-0145-FEDER-000016", financed by the North Portugal Regional Operational Programme
(NORTE 2020), under the PORTUGAL 2020 Partnership Agreement, and through
the European Regional Development Fund (ERDF). 
Quoc Dang Vu, Minh Nguyen Nhat To, Eal Kim and Jin Tae Kwak are supported by the National Research Foundation of Korea (NRF) grant funded by the Korea government (MSIP) (No. 2016R1C1B2012433).

The authors would like to thank the pathologists Dr$^{a}$ Ana Ribeiro, Dr$^{a}$ Rita Canas Marques e Dr$^{a}$ Ierec\^{e} Aymor\'{e} for their help in labeling the microscopy images.
\\\par 
The authors would also like to thank all the other BACH Challenge participants that registered, submitted their method and were accepted at the 15$^{\text{th}}$ International Conference on Image Analysis and Recognition (ICIAR 2018):
Kamyar Nazeri, Azad Aminpour, and Mehran Ebrahimi; 
Nick Weiss, Henning Kost, and Andr\'{e} Homeyer;
Alexander Rakhlin, Alexey Shvets, Vladimir Iglovikov, and Alexandr A. Kalinin;
Zeya Wang, Nanqing Dong, Wei Dai, Sean D. Rosario, and Eric P. Xing;
Carlos A. Ferreira, T\^{a}nia Melo, Patrick Sousa, Maria In\^{e}s Meyer, Elham Shakibapour and Pedro Costa;
Hongliu Cao, Simon Bernard, Laurent Heutte, and Robert Sabourin;
Ruqayya Awan, Navid Alemi Koohbanani, Muhammad Shaban, Anna Lisowska, and Nasir Rajpoot;
Sulaiman Vesal, Nishant Ravikumar, AmirAbbas Davari, Stephan Ellmann, and Andreas Maier;
Yao Guo, Huihui Dong, Fangzhou Song, Chuang Zhu, and Jun Liu;
Aditya Golatkar, Deepak Anand, and Amit Sethi;
Tomas Iesmantas and Robertas Alzbutas;
Wajahat Nawaz, Sagheer Ahmed, Ali Tahir, and Hassan Aqeel Khan;
Artem Pimkin, Gleb Makarchuk, Vladimir Kondratenko, Maxim Pisov, Egor Krivov, and Mikhail Belyaev;
Auxiliadora Sarmiento, and Irene Fond\'{o}n;
Quoc Dang Vu, Minh Nguyen Nhat To, Eal Kim, and Jin Tae Kwak;
Yeeleng S. Vang, Zhen Chen, and Xiaohui Xie;
Chao-Hui Huang, Jens Brodbeck, Nena M. Dimaano, John Kang, Belma Dogdas, Douglas Rollins, and Eric M. Gifford
(authors are ordered according to conference proceedings).





\bibliographystyle{model2-names}\biboptions{authoryear}

\input{mendeley.bbl}


\end{document}

%% file: introduction.tex
\section{Introduction}
\label{sec:intro}

Breast cancer is one of most common cancer-related death causes in women of all age~\cite{Siegel2017}, but early diagnosis and treatment can significantly prevent the disease's progression and reduce its morbidity rates~\cite{Smith2005}. Because of this, women are recommended to do self check-ups via palpation and regular screenings via ultrasound or mammography; if an abnormality is found, a breast tissue biopsy is performed~\cite{NationalBreastCancerFoundation2015}. 
Usually, the collected tissue sample is stained with hematoxylin and eosin (H\&E), which allows to distinguish the nuclei from the parenchyma, and is observed via an optic microscope. Complementarily, these samples can also be scanned to giga-pixel size images, referred as whole-slide image (WSI), for posterior digital processing. During assessment, pathologists search for signs of cancer on microscopic portions of the tissue 
by analyzing its histological properties. This procedure allows to distinguish malignant regions from non-malignant (benign) tissue, which present changes in normal structures of breast parenchyma that are not directly related with progression to malignancy. These malignant lesions can be further classified as \textit{in situ} carcinoma, where the cancerous cells are restrained inside the mammary ductal-lobular system, or invasive if the cancer cells are spread beyond the ducts.
Due to the importance of correct diagnosis in patient management the search for precise, robust and automated systems has increased. The differentiation of breast samples into normal, benign and malignant (either \textit{in situ} or invasive) brings relevant changes in the treatment of the patients making the accurate diagnosis essential. For instance, benign lesions can usually be followed clinically without the need for surgery, but malignancy almost always require surgery with or without the addition of chemotherapy.
\par
The analysis of breast cancer WSIs is non-trivial due to the large amount of data to visualize and the complexity of the task~\cite{Elmore2015}. On this setting, computer-aided diagnosis (CAD) systems can alleviate the procedure by providing a complementary and objective assessment to the pathologist. Despite the high performance of these systems for the binary classification (healthy \textit{vs} malignant) of microscopy~\cite{Kowal2013,Filipczuk2013,George2014,Belsare2015} and whole-slide images~\cite{Cruz-Roa2014,Cruz-roa2018a}, the previously referred standard clinical classification procedure has now only started to be explored~\cite{Araujo2017,Fondon2018,Han2017,Bejnordi2017}.

\subsection{Related work}

Automatic methods for breast cancer assessment in histology images can be divided according to the type of image in study (namely microscopy images and WSI) and number of classes, \textit{i.e.}, abnormality types, they consider.

\subsubsection{Microscopy images}

The classification of breast histology microscopy images as benign-malignant for referral purposes is a vastly addressed topic. Over the past decade, these methods have focused on the extraction of nuclei features, which requires the detection of these regions-of-interest. For example, nuclei have been segmented via color-based clustering \cite{Kowal2013} or by nuclei candidate detection using the circular Hough transform, followed by feature-based candidate reduction and refinement via watersheds \cite{George2014}. These segmentations allow to extract features, usually related to morphology, topology and texture. The computed features can then be used for training one or more classifiers and allow to achieve accuracies of 84-93$\%$ \cite{Kowal2013} and 72-97$\%$ \cite{George2014}.

A viable alternative to the design and extraction of hand-crafted features is to use deep learning approaches, namely convolutional neural networks (CNNs), since these allow to significatly reduce the need for field-knowledge while achieving similar or better results. For instance,
\cite{Spanhol2016} used CNNs to classify patches of microscopy images, and combined the predictions into an image label through sum, product and maximum rules. The method was evaluated on the BreaKHis dataset  \cite{Spanhol2016a} , which contains images of different magnifications, and achieved an accuracy of 84\% for a 200$\times$ magnification.

The more complex 3-class problem of considering  normal tissue, \textit{in situ} carcinoma and invasive carcinoma
has also been addressed by the scientific community. 
Due to the increased complexity of the task, using nuclei-related features is usually not sufficient to achieve a reasonable classification performance. Namely, distinguishing \textit{in situ} and invasive carcinomas requires assessing both the nuclei and their organization on the tissue. 
For instance, \cite{Zhang2011} used a cascade classification approach, where features based on the curvelet transform and local binary patterns were randomly chosen as input to a set of parallel suport-vector machines (SVMs). The images where no agreement was found were analysed by a set of NNs using another random feature set, resulting in an accuracy of 97$\%$.

Despite the successes for 2-class and 3-class classifications, few works have addressed the 4 classification problem (normal tissue, benign lesion, \textit{in situ} and invasive carcinoma) of histology images. Recently \cite{Fondon2018} proposed a handcrafted feature-based approach, considered three major sets of features, related with the nuclei, color regions and textures accounting for local and global image properties, which were then used for training a SVM. 
By their turn,  
\cite{Araujo2017} proposed a CNN-based approach, training a VGG-like network using patches extracted from the histology images. In the design of the network the authors had in consideration the effective receptive fields at each network layer in order to ensure that information is captured at different scales, allowing that both nuclei organization and the overall tissue structure could be considered. The features extracted by the CNN were then used to train a SVM, and majority voting was used for obtaining the final image label from the individual patch classifications.  
These methods have been developed in the context of the Bioimaging 2015 challenge\footnote{\url{http://www.bioimaging2015.ineb.up.pt/challenge_overview.html}}.
\cite{Fondon2018} and \cite{Araujo2017} have achieved accuracies in the 4-class problem of around 68 $\%$ and 78$\%$, respectively, on a test set of 36 images from which approximately a half correspond to extremely hard cases to classify, accordingly to two specialists.

\subsubsection{Whole-slide image analysis}

The recent advances on the acquisition systems have enabled the digitization of entire slides, avoiding loss of biopsy tissue and providing extra context for pathology assessment by the medical experts. However, automatic analysis of these images is challenging due to their giga-pixel size and wide-variety of local tissue behavior. Because of this, supervised WSI analysis has been mainly performed by assessing patches of different magnifications. For instance, the detection of invasive cancer regions can be performed by training a CNN with small patches and predicting over an entire slide. A properly trained CNN can achieve balanced accuracies (the average of specificity and sensitivity) of 84$\%$, outperforming handcrafted-feature approaches by more than 5$\%$ \cite{Cruz-Roa2014,Cruz-Roa2017}.

The majority of the solutions applied to WSIs are computationally expensive and deal only with small regions of interest (ROIs) and not the complete WSI, since this would require estimating an incredibly high number of parameters. Thus, efforts have been made for developing methods which can deal with these large sized images without the need for ROI selection or extremely high computational power while yielding a good performance.
For instance, \cite{Cruz-roa2018a} proposed the combination of CNNs and an adaptive sampling method which relies on the quasi-Monte Carlo sampling and a gradient-based adaptive strategy aiming at focusing sampling on areas of the image of higher uncertainty. The method was evaluated on 195 studies, achieving a Dice coefficient of 76 $\%$ and yielded comparable results to those from a dense sampling while greatly increasing the computational efficiency (1500$\times$ faster).

Similarly to microscopy images, the scientific community is now starting to explore multi-class classification on WSI, mainly using deep learning approaches. In
\cite{Bejnordi2017}, a context aware stack of 2 CNNs was used for detecting normal/benign tissue as well as \textit{in situ} and invasive carcinomas. The first CNN was trained for classifying small high-resolution patches of the WSIs, thus learning celular-level characteristics. This fully-convolutional model was then used for predicting a set of feature maps from patches of higher size, which serve as input for a second CNN. This scheme allows to integrate both local and global features related to tissue organization, achieving accuracies of 82$\%$ and a Cohen's kappa value of 0.7.

By its turn, \cite{Gecer2018} considered 5 classes for the detection and classification of cancer in WSIs: non proliferative changes only, proliferative changes, atypical ductal hyperplasia, \textit{in situ} and invasive carcinoma. The classification was performed by combining the prediction of two steps. First, four CNNs were used sequentially to detect ROIs in a multiscale fashion. Specifically, the output of each CNN is a set of ROIs for the next, thus increasing the magnification of the analyzed region. Then, the output of the fourth model is classified by a CNN trained with high magnification patches. The outputs of the classification and ROI-proposal CNNs are then combined via majority voting, allowing to obtain a slide-level accuracy of 55$\%$, which revealed no statistical difference from 45 pathologists' performance. 

\subsection{Challenges}

Challenges are known to enable advances on the medical image analysis field by promoting the participation of multiple researchers of different backgrounds on a competitive, but scientifically constructive, setting. Over the past years, the scientific community has been promoting challenges on different imaging modalities and topics. Related to breast cancer, CAMELYON \footnote{\url{https://camelyon17.grand-challenge.org/}} is a two-edition challenge aimed at cancer metastases detection on WSI of lymph node sections.

To further promote and complement the research on the breast cancer image analysis field, the Grand Challenge on BreAst Cancer Histology images (BACH) was organized as part of the ICIAR 2018 conference (15th International Conference on Image Analysis and Recognition)\footnote{\url{https://www.aimiconf.org/iciar18/}}. BACH is a biomedical image challenge built on top of the Bioimaging 2015 challenge, with a much larger dataset of H\&E stained microscopy images for classification and a new set of WSI breast cancer tissue images for segmentation. Specifically, the participants of BACH were asked to predict the type of these tissue samples as
\begin{inparaenum}[1)]
\item Normal,
\item Benign
\item \textit{In~situ} carcinoma and
\item Invasive carcinoma,
\end{inparaenum}
 with the goal of providing pathologists a tool to reduce the diagnosis workload. The rest of the paper is organized as follows.
Section~\ref{sec:challenge} details the challenge in terms of organization, dataset and participant's evaluation. Then, Section~\ref{sec:methods} describes the approaches of the best performing methods, and the corresponding performance is provided on Section~\ref{sec:results}. Finally, Section~\ref{sec:conclusion} summarizes the findings of this study.

%% file: materials.tex
\section{Challenge description}
\label{sec:challenge}

\subsection{Organization}

\begin{figure*}[t]
\centering
\includegraphics[width=\textwidth]{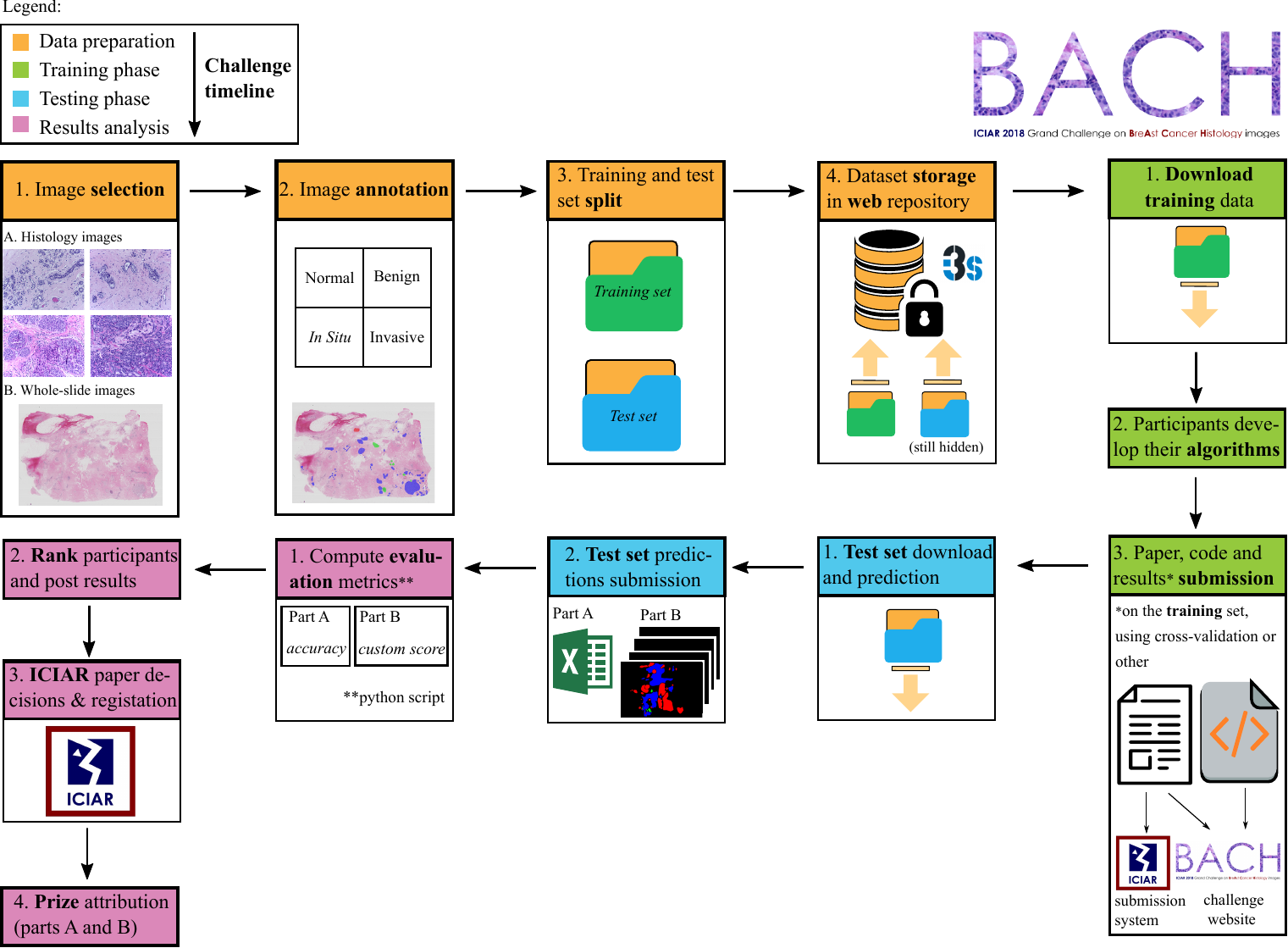}
\caption{Workflow of the BACH challenge.} 
\label{fig:challenge_overal}
\end{figure*}

The BACH challenge was organized into different stages, providing a well structured workflow to potentiate the success of the initiative (Fig.~\ref{fig:challenge_overal}).
The challenge was hosted on \textit{Grand Challenge}\footnote{\url{https://grand-challenge.org/}}, which allowed for an easy platform set-up. At the time of this writing, \textit{Grand Challenge} accounts for more than 12~000 registered users and, alongside Kaggle\footnote{\url{https://www.kaggle.com/}}, it is one of the preferred platforms for medical imaging-related  challenges. The BACH was also announced via the \textit{Sci-diku-imageworld} mailing list\footnote{\url{https://list.ku.dk/listinfo/sci-diku-imageworld}}. 
Participants were asked to register on the \textit{Grand Challenge} to access most of the contents of the BACH webpage. All registrations were manually validated by the organization to minimize spam and anonymous participation. Once accepted, participants could download the data by filling a form asking for their name, institution and e-mail address. Once the form was submitted, an e-mail containing an unique set of credentials (username, password) and the dataset download link was automatically sent to the provided e-mail address. This allowed the organization to better limit the dataset access to non-participants as well as collect a list of the institutions/companies interested in the challenge.
\par
BACH was divided in two parts, \textbf{A} and \textbf{B}. \pA{} consisted in automatically classifying H\&E stained breast histology microscopy images in four classes: 
\begin{inparaenum}[1)]
\item Normal,
\item Benign,
\item \textit{In situ} carcinoma and
\item Invasive carcinoma.
\end{inparaenum}
\pB{} consisted in performing pixel-wise labeling of whole-slide breast histology images in the same four classes.
Participants were allowed to participate on a single part of the challenge.  Also, to promote participation (thus more competition and higher quality of the methods), the ICIAR 2018 conference sponsored the challenge by awarding monetary prizes to the first and second best performing methods, for both challenge parts.
The prize awarding was contingent of the acceptance and presentation of the methodology at the ICIAR 2018 conference.   

\par
The BACH website\footnote{\url{https://iciar2018-challenge.grand-challenge.org/}} was first made publicly available on the $1^{st}$ November 2017 with the release of the labeled training set. The registered participants had up to $1^{st}$ February 2018 (4 months) to submit the source code of their methods and a paper describing their approach. To promote the dissemination of the methods, participants were also required to submit their paper to the ICIAR conference. The test set was released on the $5^{th}$ February 2018 and submissions were open for a week. Results were announced a month after, at the $12^{th}$ March 2018.

\subsection{Datasets}

The BACH challenge made available two labeled training datasets for the registered participants. The first dataset is composed of microscopy images annotated image-wise by two expert pathologists from the Institute of Molecular Pathology and Immunology of the University of Porto (IPATIMUP) and from the Institute for Research and Innovation in Health (i3S). The second dataset contains pixel-wise annotated and non-annotated WSI images. For the WSI, annotations were performed by a pathologist and revised by a second expert. The training data is publicly available at \url{https://iciar2018-challenge.grand-challenge.org/}.

\subsubsection{Microscopy images dataset}

\begin{figure*}[tb]

\begin{subfigure}{0.22\textwidth}
\includegraphics[width=\textwidth]{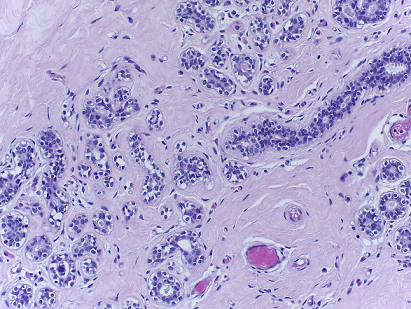}
\caption{Normal}
\end{subfigure}
\hfill
\begin{subfigure}{0.22\textwidth}
\includegraphics[width=\textwidth]{figures/benign}
\caption{Benign}
\end{subfigure}
\hfill
\begin{subfigure}{0.22\textwidth}
\includegraphics[width=\textwidth]{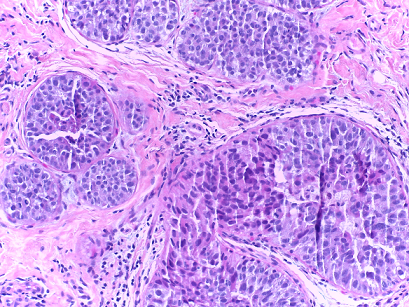}
\caption{\textit{In situ}}
\end{subfigure}
\hfill
\begin{subfigure}{0.22\textwidth}
\includegraphics[width=\textwidth]{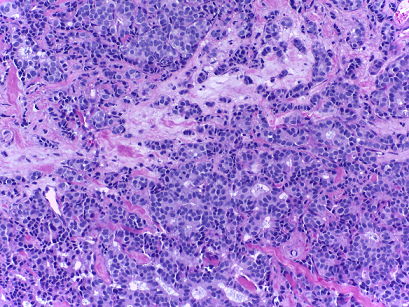}
\caption{Invasive}
\end{subfigure}
\caption{Examples of microscopy images from the BACH dataset.}
\label{fig:micros}
\end{figure*}

The microscopy dataset is composed of 400 training and 100 test images, with the four classes equally represented (see Fig.~\ref{fig:micros}).
All images were acquired in 2014, 2015 and 2017 using a Leica DM 2000 LED microscope and a Leica ICC50 HD camera and all patients are from the Porto and Castelo Branco regions (Portugal). Cases are from Ipatimup Diagnostics and come from three different hospitals (Hospital CUF Porto, Centro Hospitalar do T\^{a}mega e Sousa and Centro Hospitalar Cova da Beira). The annotation was performed by two medical experts. 
Images where there was disagreement between the Normal and Benign classes were discarded. The remaining doubtful cases were confirmed via imunohistochemical analysis. The provided images are on RGB \texttt{.tiff} format and have a size of $2048 \times 1536$ pixels and a pixel scale of 0.42 $\mu$m $\times$ 0.42 $\mu$m. The labels of the images were provided in \texttt{.csv} format.
Participants were provided with a partial patient-wise distribution of the images of the training set. The test data was collected from a completely different set of patients, ensuring a fairer evaluation of the methods. Note that the training set is an extension of the one used for developing the approach in~\cite{Araujo2017}.

\subsubsection{Whole-slide images dataset}

\begin{figure}
\centering
\includegraphics[width=\columnwidth]{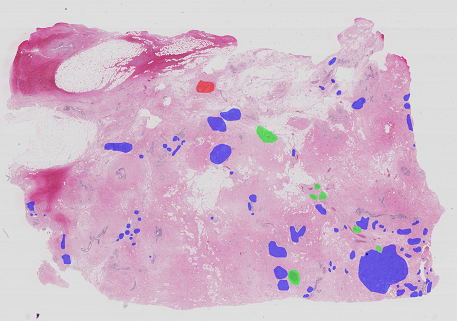}
\caption{Example of a pixel-wise annotated whole-slide image from the training set . {\color{red}$\blacksquare$}$~$benign; {\color{green}$\blacksquare$}$~$\textit{in situ}; {\color{blue}$\blacksquare$}$~$invasive. \label{fig:wsi}}
\end{figure}

Whole-slide images (WSI) are high resolution images containing the entire sampled tissue. 
Because of that, each WSI can have multiple pathological regions. The BACH's \pB{} dataset is composed of 30 WSI for training and 10 WSI for algorithm testing. Specifically for training, the organization provided 10 pixel-wise annotated regions for the Benign, \textit{In situ} carcinoma and Invasive carcinoma classes and 20 potentially pathological WSIs that were not annotated by the experts. The provided annotations aim at identifying regions of interest for the diagnosis on the lowest magnification setting and thus may include non-tissue and normal tissue regions, as depicted in Fig.~\ref{fig:wsi}. The distribution of the labels is shown in Table~\ref{tab:labels_pB}.

The WSI images were acquired in 2013–2015 from patients from the Castelo Branco region (Portugal) with a Leica SCN400 (from Centro Hospitalar Cova da Beira), and were made available on \textit{.svs} format, with a  pixel scale of 0.467 $\mu$m/pixel and variable size with width~$\in$~[39980~62952] and height~$\in$~[27972~44889] (pixels). The ground-truth was released as the coordinates of the points that enclose each labeled region via a \texttt{.xml} file.

\begin{table}
\centering
\caption{Relative distribution (\%) of the labels for the training and test sets of \pB{}. \label{tab:labels_pB}}
\begin{tabular}{cccc}\hline

&\textbf{Benign}&\textit{\textbf{In situ}}& \textbf{Invasive} \\\hline
\textbf{Train} & 9 & 3 & 88 \\
\textbf{Test} & 31 & 6 & 63\\\hline

\end{tabular}
\end{table}

\subsection{Performance Evaluation}

The methods developed by the participants were evaluated on independent test sets for which the ground-truth was hidden. Specifically, for \pA{} participants were requested to submit a \texttt{.csv} containing row-wise pairs of (image name, predicted label) for the 100 microscopy images. 
Performance on the microscopy images was evaluated based on the overall prediction accuracy, \textit{i.e.}, the ratio between correct samples and the total number of evaluated images.

For \pB{} it was required the submission of $4\times$~downsampled WSI \texttt{.png} masks with values 0~--~Normal, 1~--~Benign, 2~--~\textit{In situ} carcinoma and 3~--~Invasive carcinoma. Possible mismatches between the prediction's and ground truth's sizes were corrected by padding or cropping the prediction masks.    
The performance on the WSI images was evaluated based on the custom score $s$:
\begin{equation}
s = 1-\frac{\sum_{i=1}^N{|\texttt{pred}_i-\texttt{gt}_i|}}{\displaystyle\sum_{i=1}^N max(\texttt{gt}_i,|\texttt{gt}_i-3|) \times [1-(1-\texttt{pred}_{i,bin})(1-\texttt{gt}_{i,bin})]+a}
\label{eq:score}
\end{equation}
where $\texttt{pred}$ is the predicted class (0, 1, 2 or 3), and $\texttt{gt}$ is the ground truth class, $i$ is the linear index of a pixel in the image,  $N$ is the total number of pixels in the image, $X_{bin}$ is the binarized value of $X$, \textit{i.e.}, is 0 if the label is 0 and 1 otherwise, and $a$ is a very small number that avoids division by zero. 
Fig.~\ref{fig:score_examples} illustrates the results of this metric on a set of predictions which are gradually farther from the ground truth.

This score is based on the accuracy metric, aiming at penalizing more the predictions that are farther from the ground truth value. The reasoning behind this metric is the following: in the numerator, the absolute distance between the predicted class and the ground truth is measured for all samples $i$, which is indicative of how far the prediction is from the ground truth, \textit{e.g.}, if $\texttt{gt} = 1$ and $\texttt{pred} = 3$, the distance is 2, whereas if $\texttt{pred} = 2$ the distance is 1. To normalize these distances, in the first factor of the denominator we consider, for each sample, the largest distance possible having in account the true label, \textit{e.g.}, if $\texttt{gt} = 0$ the maximum distance possible is 3 ($max(0,|0-3|)=3$), while if $\texttt{gt} = 2$ the maximum distance is 2 ($max(2,|2-3|)=2$).
Also, the cases in which the prediction and ground truth are both 0 (Normal class) are not counted, since these can be seen as true negative cases. This is allowed by the second factor of the multiplication of the denominator, which is equal to zero if $\texttt{pred} = 0$ and $\texttt{gt} = 0$, and equal to 1 otherwise.

\begin{center}
    \centering
    \includegraphics[width=\columnwidth]{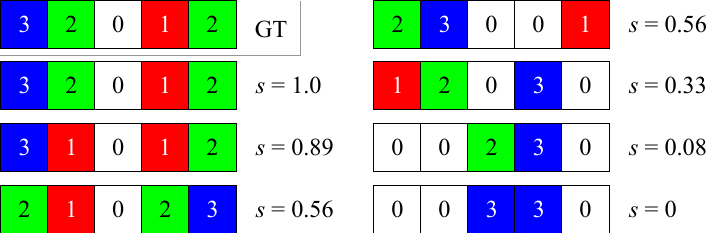}
    \captionof{figure}{Examples of the custom score metric.
    {$\square$}\,$|$\,0$~$normal; 
    {\color{red}$\blacksquare$}\,$|$\,1$~$benign; 
    {\color{green}$\blacksquare$}\,$|$\,2$~$\textit{in$~$situ}; 
    {\color{blue}$\blacksquare$}\,$|$\,3$~$invasive.
    \label{fig:score_examples}}   
\end{center}

Note that this custom evaluation score was preferred over both the Intersection over Union (IoU) and the quadratic weighted Cohen’s Kappa statistic~\cite{Cohen1960}. Namely, the custom score allows to ignore correct Normal class predictions (highly dominant) while penalizing wrong Normal predictions, whereas for Kappa the Normal class would have either to be completely considered or ignored. 
Likewise, 
the custom metric is not only able to assess if the methods were properly capable of detecting pathological regions on whole-slide images (as would the IoU, if applied class-wise), but also of indicating how far that prediction is from the ground-truth. 
Given the complexity of the task, the direct computation of the IoU could over-penalize methods that are capable of finding abnormal regions but fail to correctly classify them. By assessing the distance of the predictions, the custom metric has a higher clinical relevance than analyzing region overlaps. For instance, mispredicting Normal tissue as Benign should be considered less severe than predicting it as Invasive.

%% file: methods.tex
\section{Competing solutions} 
\label{sec:methods}

This section provides a comprehensive description of the participating approaches. 
Table~\ref{tab:partA} and Table~\ref{tab:partB} summarize the methods that achieved an accuracy $\geq 0.7$ and score $\geq 0.5$ on \pAB{}, respectively. The most relevant methods in terms of performance and applicability are detailed on the next sections. For methods that solve \pAB{} jointly, refer to Section~\ref{sec:partAB}, and for \pA{} or \pB{} exclusively refer to Sections~\ref{sec:partA}~and~\ref{sec:partB}, respectively.

\input{table_partA}

\input{table_partB}

\subsection{Introduction to Convolutional Neural Networks}

The vast majority of \pA{} and all of \pB{} participants proposed a convolutional neural network (CNN) approach to solve BACH. CNNs are now the state-of-the-art approach for computer vision problems and show high promise in the field of medical image analysis \cite{Litjens2017,Tajbakhsh2016} because they are easy to set up, require little applied field knowledge (specially when compared with handcrafted feature approaches) and allow to migrate base features from generic natural image applications~\cite{Deng2009}. 
\par
CNN performance is highly dependent on the architecture of the network as well as on the hyper-parameter optimization (learning rate, for instance). The large number of parameters in CNNs make them prone to overfit to the training data, specially when a relatively low number of training images is available. Because of that, it is a common practice in medical image analysis to fine-tune networks trained on medical images. In BACH, participants opted mainly for pre-trained networks that have historically achieved high performance in the ImageNet natural image analysis challenge~\cite{Russakovsky2015}. From those, VGG~\cite{Simonyan2014}, Inception~\cite{Szegedy2015}, ResNet~\cite{He2016} and DenseNet~\cite{Huang2017} were the ones that achieved the overall higher results. A brief description of these networks is provided bellow.
\par

VGG (Visual Geometry Group) was one of the first networks to show that increasing model depth allows higher prediction performance. This network is composed of blocks of 2-3 convolutional layers with a large number of filters that are followed by a max pooling layer. The output of the last layer is then connected to a set of fully-connected layers to produce the final classification. However, despite the success of this model on the ImageNet challenge, the linear structure of VGG and large number of parameters (approximately 140M for 16 layers) does not allow to significantly increase the depth of the model and increases tendency to overfit.
\par
The Inception network follows the theory that most activations in a (deep-)CNN are either unnecessary or redundant and thus the number of parameters can be reduced by using locally sparse building blocks (\textit{a.k.a.} inception blocks). At each inception block, the number feature maps of the previous block is reduced via an $1\times 1$ convolution. The projected features are then convolved in parallel by kernels of increasing size, allowing to combine information at multiple scales. Finally, replacing the the fully-connected layers by a global average pooling allows to significantly reduce the model parameters (23M parameters with 159 layers) and makes the network fully convolutional, enabling its application to different input sizes.
\par
Increasing network depth leads to vanishing gradient problems as a result of the large number of multiplication operations. Consequently, the error gradient will be vanishingly small preventing effective updates of the weights in the initial layers of the model. The recent versions of Inception tackle this issue by using Batch Normalization \cite{Ioffe2015}, which allows to reestablish the gradient by normalizing the intermediary activation maps with the statistics of the training batch. Alternatively, ResNet uses residual blocks to stabilize the value of the error gradient during training. In each residual block the input activation map is summed to the output of a set of convolutional layers, thus stopping the gradient from vanishing and easing the flow of information. A ResNet with 50 residual blocks (169 layers) has approximately 25M parameters.
\par
The high performance of models like Inception or ResNet have strengthened the deep learning design principle that "deeper networks are better" by improving on the feature redundancy and gradient vanishing problems. Recently, an even deeper network, DenseNet, has addressed these same issues by using dense blocks. Dense blocks introduce short connections between convolutional layers, \textit{i.e.}, for each layer, the activations of all preceding layers are used as inputs. By doing so, DenseNet promotes feature re-use, reducing the feature redundancy and thus allowing to decrease the number of feature maps per layer. Specifically, a DenseNet with 201 layers has approximately 20M parameters to optimize.

As already mentioned, fine-tuning of high performance networks trained in natural images is the preferred approach for medical image analysis. Fine-tuning for classification is usually performed as follows:
\begin{inparaenum}[1)]
\item the network is initialized with weights trained to solve a natural image classification problem such as the ImageNet classification task;
\item the classification head, usually a fully-connected layer, is replaced by a new one with randomly initialized parameters;
\item initially, the new classification head is trained for a fixed number of iterations by inputting the medical images and inhibiting the filters of the pre-trained model to change;
\item then, different blocks of the pre-trained model are progressively allowed to learn and adapt to the new features, allowing the model to move to new local optima and increase the overall performance of the network.
\end{inparaenum}
\par

\subsection{\pA{} \label{sec:partA}}


\subsubsection{\etal{Chennamsetty} (team 216)}


\cite{Chennamsetty2018} used an ensemble of ImageNet pre-trained CNNs to classify the images from \pA{}. Specifically, the algorithm is composed of a ResNet-101\cite{He2016} and two DenseNet-161~\cite{Huang2017} networks fine-tunned with images from varying data normalization schemes. 
Initializing the model with pre-trained weights alleviates the problem of training the networks with limited amount of high quality labeled data.
First, the images were resized to $224\times224$ pixels via bilinear interpolation and normalized to zero mean and unit standard deviation according to statistics derived either from ImageNet or \pA{} datasets, as detailed below.
\par
During training, the ResNet-101 and a DenseNet-161 were fine-tuned with images normalized from the breast histology data whereas the other DenseNet-161 was fine-tuned with the ImageNet normalization. Then, for inference, each model in the ensemble predicts the cancer grade in the input image and a  majority voting scheme is posteriorly used for assigning the class associated with the input.

\subsubsection{\etal{Brancati} (team 1)}


\cite{Brancati2018} proposed a deep learning approach based on a fine-tuning strategy by exploiting transfer learning on an ensemble of ResNet~\cite{He2016} models. ResNet was preferred to other deep network architectures because it has a small number of parameters and shows a relatively low complexity in comparison to other models.
The authors opted by further reducing the complexity of the problem by down-sampling the image by factor $k$ and using only the central patch of size $m\times m$ as input to the network. In particular, $k$ was fixed to 80\% of the original image size and $m$ was set equal to the minimum size between the width and high of the resized image. 
\par
The proposed ensemble is composed of 3 ResNet configurations: 34, 50 and 101. Each configuration was trained on the images from \pA{} and the classification of a test image is obtained by computing the highest class probability provided by the three configurations.

\subsubsection{\etal{Wang} (team 157)}


\cite{Wang2018} proposed the direct application of VGG-16~\cite{Simonyan2014} to solve \pA{}. Prior to fine-tuning the model, all images from \pA{} were resized to $256\times 256$ and normalized to zero mean and unit standard deviation. To account for the model input size, training is performed by cropping patches of $224\times 224$ pixels at random locations of the input image. 
First, the model is trained using a Sample Pairing~\cite{Inoue2018} data augmentation scheme. 
Specifically, a random pair of images of different labels is independently augmented (translations, rotations, etc.) and then superimposed with each other. 
The resulting mixed patch receives the label of one of the initial images and is afterwards used to train the classifier. 
The learned weights are then used as a starting point to train the network with the initial (\textit{i.e.} non mixed) dataset.

\subsubsection{Kone \textit{et. al} (team 19)}


\cite{Kone2018} proposed a hierarchy of 3 ResNeXt50~\cite{Xie2017} models in a binary tree like structure (one parent and two child nodes) for the 4-class classification of \pA{}. 
The top CNN classifies images in two high level groups: 
\begin{inparaenum}
\item carcinoma, which includes the \textit{in situ} and the invasive classes and 
\item non-carcinoma, which includes normal and benign.
\end{inparaenum}
Then, each of children CNNs sub-classifies the images in the respective 2 classes.
\par
The training is performed in two steps. First, the parent ResNeXt50 pre-trained on ImageNet is fine tunned with the images from \pA{}. The learned filters are then used as the starting point for the child networks. The authors also divide the ResNeXt50 layers into three groups and assign them different learning rates based on the optimal one found during training.

\subsection{\pB{} \label{sec:partB}}




\subsubsection{\etal{Galal} (team 264)}

\cite{Galal2018} proposed Candy Cane, a fully convolutional network based on DenseNets~\cite{Huang2017} for the segmentation of WSIs. Candy Cane was designed following an auto-encoder scheme of downsampling and upsampling paths with skip connections between corresponding down and up feature maps to preserve low level feature information. Accounting for GPU memory restrictions, the authors propose a downsampling path much longer than the upsampling counter-part. Specifically, Candy Cane operates on $2048\times 2048$ slice images and outputs the corresponding labels at a size of $256\times256$ pixels. Similarly to an expert that looks at a microscope in few adjacent regions to examine the tissue but then identifies regions in the larger context of the tissue, the large input size of the model allows the network to have both microscopy and tissue organization contexts. The output of the system is then resized to the original size.


\subsection{\pAB{} \label{sec:partAB}}

\subsubsection{\etal{Kwok} (team 248)}




\cite{Kwok2018} used a two-stage approach to take advantage of both microscopy and WSI images. 
To account for the partially missing patient-wise origin on \pA{}, images which origin was not available were clustered based on color similarity. The data was then split accordingly.
For stage 1, $5600$ patches of $1495\times1495$ pixels were extracted from \pA's images with a stride of 99 pixels.
These patches were then resized to $299\times299$ pixels and used for fine-tuning a 4 class Inception-Resnet-v2~\cite{Szegedy2017} trained on ImageNet. 
This network was then used for analyzing the WSIs. Specifically, WSI foreground masks were computed via a threshold on the L*a*b color space. 
Then, patches were extracted from the WSIs in the same way as for \pA{}. This second patch dataset was refined by discarding images with $<5\%$ foreground and posteriorly labeled using the CNN trained on \pA. Finally, $5900$ patches from the top 40\% incorrect predictions (evenly sampled from each of the 4 classes) were selected as hard examples for stage 2.
\par
For stage 2, the CNN was retrained by combing the patches extracted from \pA{} ($5600$) and \pB{} ($5900$). The resulting model was used for labeling both microscopy images and WSIs. Prediction results were aggregated from patch-wise predictions back onto image-wise predictions (for \pA{}) and WSI-wise heatmaps (for \pB{}). 
Specifically for \pB{}, the patch-wise predictions were mapped to hard labels (Normal=0, Benign=1, \textit{in situ}=2 and Invasive=3) and combined into a single image based on the patch coordinates and the network's stride. The resulting map was then normalized to $[0\,1]$ and multi-thresholded at $\{0, 0.35,$ $0.7, 0.75\}$ to bias the predictions more towards Normal/Benign and less to \textit{in situ} and Invasive carcinomas.

\subsubsection{\etal{Marami} (team 16)}


\cite{Marami2018} proposed a classification scheme based on an ensemble of four  modified Inception-v3~\cite{Szegedy2016} CNNs that aims at increasing the generalization capability of the method by combining different networks trained on random subsets of the data. Specifically, the networks were adapted by adding an adaptive pooling before a set of custom fully connected layers, allowing higher robustness to small scale changes.
Each of these CNNs was trained via a 4 fold cross-validation approach on 512$\times$512 images extracted at 20$\times$ magnification from both microscopy images from \pA{} and WSI from \pB{}, as well as with benign tissue images from the BreakHis public dataset~\cite{Spanhol2016}.
\par
Predictions on unseen data are inferred by averaging the output probabilities of the trained ensemble network for each class, making the system more robust to potential inconsistencies and corruption in the labeled data. For \pA{}, the final label was obtained by majority voting of 12 overlapping 512$\times$512 regions. For the WSIs, local predictions were generated by using a 512$\times$512 sliding window of stride 256 pixels. The resulting output map was then refined using a ResNet34\cite{He2016} to separate tissue regions from the background and regions with artifacts, reducing potential misclassifications due to ink and other artifacts in whole slide images.

\subsubsection{\etal{Kohl} (team 54)}







\cite{Kohl2018} used an ImageNet pretrained DenseNet~\cite{Huang2017} to approach both parts of the challenge.
For \pA{}, the 400 training images were downsampled by a factor 10 and normalized to zero mean and unit standard deviation. The network was then trained on two steps:
\begin{inparaenum}[1)]
\item fine-tuning the fully-connected portion of the network by 25 epochs to avoid over-fitting and
\item training the entire network for 250 epochs.
\end{inparaenum}
\par
For \pB{}, the authors extracted patches of $330{\mu m} \times 330{\mu m}$ ($157 \times 157$ pixels) from the annotated WSIs. This patch dataset was then refined by removing patches consisting of at least 80\% background pixels, similarly to \etal{Litjens}~\cite{Litjens2016}). Due to the very limited amount of data in the benign and \textit{in situ} carcinoma classes, the authors did not perform WSI-wise split for validation purposes and instead used a randomly split dataset. Also, 16 of the 20 originally non-annotated WSIs were also annotated with the help of a trained pathologist and thus in total $41506$ image patches ($25230$ normal, $1723$ benign, $1759$ \textit{in situ} and $12794$ invasive carcinomas) were used. Network training was similar for \pA{} and \pB{}: 25 epochs for training the fully-connected layers followed by 250 epochs for training the whole network in case of \pA{}, and 6 epochs for training the fully-connected layers followed by 100 epochs for training the whole network using log-balanced class weights in case of \pB{}.

\subsubsection{\etal{Vu} (team 166)}


\cite{Vu2018} proposed to use an encoder-decoder network to solve both \pA{} and \pB{}. For \pA{}, the authors use the encoder part of the model. The encoder is composed of five convolutional processing blocks that integrate dense skip connections, group and dilated convolutions, and self-attention mechanism for dynamic channel selection following the design trends of DenseNet, Squeeze-Excitation Network (SENet) and ResNext~\cite{Huang2017,Jegou2016,Yu2015,Hu2017,Xie2017}. For classifying the microscopy images, the model has a head composed of a global average pooling and a fully-connected softmax layer. Training is performed by downsampling the images $4\times$ and online data augmentation is used.
\par
For \pB{} the full encoder-decoder scheme is used. This segmentation network follows the U-Net~\cite{Ronneberger2015} structure with skip connections between the downsample and upsample, the decoder is composed by the same convolutional blocks and the upsample is performed considering the nearest neighbor. Also, to ease network convergence, the encoder is initialized with the weights learned from \pA{}. 
For training, the WSI are first downscaled by a factor of 4 and sub-regions containing the labels of interest are collected. Specifically, the authors collect $6129$ sub-regions of size $1000\times 1000$ from which the central regions of $630\times 630$ are used as input to the model. The corresponding output segmentation map has a size of $204\times 204$. To prioritize the detection of pathological regions, the segmentation network is trained with two categorical crossentropy loss terms, where the main loss targets for the four histology classes and the auxiliary loss is computed for normal and benign  \textit{vs} \textit{in situ} carcinoma and invasive carcinoma groups.

%% file: table_partA.tex
\setlength{\tabcolsep}{0.25em} 
\renewcommand{\arraystretch}{1}

\begin{table*}
\centering
\footnotesize
\caption[Summary of the methods submitted for \pA{}.]{Summary of the methods submitted for \pA{}.
\upper{A}detailed description in Section~\ref{sec:partA};
\upper{AB}detailed description in Section~\ref{sec:partAB}. Pre-training is performed on ImageNet~\cite{Krizhevsky2012}. \textbf{Acc.} is the overall prediction success for the four classes; \textbf{Approach} lists the main methods to label the images; \textbf{Ensemble} (Ens.) indicates if the approach uses a single or multiple models (and their number, when available); \textbf{External sets} indicates if the method was trained using datasets other than from \pA{}; \textbf{Context (area ratio)} is the ratio between the original size and the size of the patch used for training the network (prior to rescaling); \textbf{Input size (pixels)} is the size of the image to be analyzed by the model; \textbf{Color normalization} (color norm.) indicates if any histology-inspired normalization was used. N/A: information not available. $^1$Pre-trained on CAMELYON (\url{https://camelyon17.grand-challenge.org/}). M1: \cite{Bejnordi2016}; M2: \cite{Krishnan2012}; M3: \cite{Macenko2009}; M4: \cite{Reinhard2001}.
\label{tab:partA}}
\begin{tabular}{lllcccccc}

\hline

\multicolumn{1}{c}{\textbf{Team}} & \multicolumn{1}{c}{\textbf{Acc.}} & \multicolumn{1}{c}{\textbf{Approach}}  & \parbox{3em}{\centering\textbf{Pre- \\ trained}} & \textbf{Ens.}  & \textbf{External sets}                                                                                               & \parbox{1.5cm}{\centering\textbf{Context \\ (area ratio)}} & \parbox{1.5cm}{\centering\textbf{Input size \\ (pixels)}} & \parbox{1cm}{\centering\textbf{Color \\ norm.}}\\ \hline 

\cite{Chennamsetty2018} (216)\upper{A}                          & 0.87                                        & \begin{tabular}[c]{@{}l@{}}Resnet-101; Densenet-161\end{tabular}                                                                 & \cmark                     & 3                                                                                & \xmark                                                                                                                     & 1                                                     & 224$\times$224                                                          & \xmark                                                                                                                                                                                                                                                                                                                                                                      \\\hline
\cite{Kwok2018} (248)\upper{AB}                             & 0.87                                        & Inception-Resnet-v2                                                                                                                & \cmark                     & \xmark                                                                           & \pB{}                                                                                                                   & 0.71                                                  & 299$\times$299                                                          & \xmark                                                                                                                                                                                                                                                                                                                                                                      \\\hline
\cite{Brancati2018} (1)\upper{A}                               & 0.86                                        & Resnet-34, 50, 101                                                                                                                 & \cmark                     & 3                                                                                & \xmark                                                                                                                     & 1                                                     & \begin{tabular}[c]{@{}c@{}}308$\times$308\\ 615$\times$615\end{tabular} & \xmark                                                                                                                                                                                                                                                                                                                                                                      \\\hline
\cite{Marami2018} (16)\upper{AB}                             & 0.84                                        & Inception-v3                                                                                                                       & \cmark                     & 4                                                                                & \begin{tabular}[c]{@{}c@{}}\pB{} \\ BreakHis\end{tabular}                                                               & 0.33                                                 & 512$\times$512                                                          & M1                                                                                                                                                                                                                                                                                                                                                                      \\\hline
\cite{Kohl2018} (54)\upper{AB}                              & 0.83                                        & Densenet-161                                                                                                                       & \cmark $^1$                    & \xmark                                                                           & \xmark                                                                                                                     & 1                                                     & 205$\times$154                                                          & \xmark                                                                                                                                                                                                                                                                                                                                                                      \\\hline
\cite{Wang2018} (157)\upper{A}                             & 0.83                                        & VGG16                                                                                                                              & \cmark                     & \xmark                                                                           & \xmark                                                                                                                     & 0.765                                                 & 224$\times$224                                                          & \xmark                                                                                                                                                                                                                                                                                                                                                                      \\\hline
\etal{Steinfeldt} (186)                                     & 0.81                                        & \begin{tabular}[c]{@{}l@{}}XCeption\end{tabular}                                              & \cmark                     & \xmark                                                                           & \xmark                                                                                                                     & 0.028-0.751                                           & 229$\times$229                                                          & \xmark                                                                                                                                                                                                                                                                                                                                                                      \\\hline
\cite{Kone2018} (19)\upper{A}                              & 0.81                                        & ResNeXt50                                                                                                                          & \cmark                     & \cmark                                                                           & BISQUE                                                                                                                     & 1                                                     & 299$\times$299                                                          & \xmark                                                                                                                                                                                                                                                                                                                                                                      \\\hline
\etal{Nedjar} (36)                                      & 0.81                                        & \begin{tabular}[c]{@{}l@{}}Inception-v3, Resnet-50, MobileNet\end{tabular}                                                    & \cmark                     & \cmark                                                                           & \xmark                                                                                                                     & 1                                                     & 224$\times$224                                                          & \xmark                                                                                                                                                                                                                                                                                                                                                                      \\\hline
\etal{Ravi} (412)                                     & 0.8                                         & Resnet-152                                                                                                                         & \cmark                     & \xmark                                                                           & \xmark                                                                                                                     & 0.875                                                 & 224$\times$224                                                          & M2                                                                                                                                                                                                                                                                                                                                                                      \\\hline
\cite{ZWang2018} (22)                                       & 0.79                                        & VGG16                                                                                                                              & \cmark                     & \xmark                                                                           & \xmark                                                                                                                     & 0.255                                                 & 224$\times$224                                                          & M3                                                                                                                                                                                                                                                                                                                                                                      \\\hline
\cite{Cao2018} (425)                                      & 0.79                                        & \begin{tabular}[c]{@{}l@{}}PTFAS+GLCM, ResNet-18, ResNeXt,\\ NASNet-A, ResNet-152, VGG16,\\ Random Forest SVM\end{tabular} & \cmark                     & \begin{tabular}[c]{@{}c@{}}\cmark\\ \end{tabular} & \xmark                                                                                                                     & 1                                                     & \begin{tabular}[c]{@{}c@{}}224$\times$224\\ 331$\times$331\end{tabular} & \xmark                                                                                                                                                                                                                                                                                                                                                                      \\\hline
\etal{Seo} (60)                                      & 0.79                                        & \begin{tabular}[c]{@{}l@{}}ResNet, Inception-V3, Random Forests\end{tabular}                                                     & \cmark                     & \xmark                                                                           & \begin{tabular}[c]{@{}c@{}}\cmark\\ \end{tabular} & 1                                                     & 299$\times$299                                                          & \xmark                                                                                                                                                                                                                                                                                                                                                                      \\\hline
\etal{Sidhom} (370)                                     & 0.78                                        & ResNet-50                                                                                                                          & \cmark                     & \xmark                                                                           & \xmark                                                                                                                     & \begin{tabular}[c]{@{}c@{}}0.018\\ 0.289\end{tabular} & 224$\times$224                                                          & \begin{tabular}[c]{@{}c@{}}M4\\ \\ \end{tabular}                                                                                                                                               \\\hline
\cite{Guo2018} (242)                                      & 0.77                                        & GoogLeNet                                                                                                                          & \cmark                     & 2                                                                                & \xmark                                                                                                                     & \begin{tabular}[c]{@{}c@{}}1\\ 0.083\end{tabular}     & 224$\times$224                                                          & \begin{tabular}[c]{@{}c@{}} M4\\ \\ \end{tabular}                                                                                                                                               \\ \hline
\etal{Ranjan} (61)                                      & 0.77                                        & AlexNet                                                                                                                            & \cmark                     & 2                                                                                & \xmark                                                                                                                     & 1                                                     & 224$\times$224                                                          & \xmark                                                                                                                                                                                                                                                                                                                                                                      \\\hline
\cite{Mahbod2018} (73)                                       & 0.77                                        & \begin{tabular}[c]{@{}l@{}}ResNet-50, ResNet-101\end{tabular}                                                                     & \cmark                     & 2                                                                                & \xmark                                                                                                                     & 1                                                     & 224$\times$224                                                          & \begin{tabular}[c]{@{}c@{}} M3\\ \end{tabular}          \\\hline

\cite{Ferreira2018} (18)                                       & 0.76                                        & Inception-ResNet-v2                                                                                                                & \cmark                     & \xmark                                                                           & \xmark                                                                                                                     & 1                                                     & 224$\times$224                                                          & \xmark                                                                                                                                                                                                                                                                                                                                                                      \\\hline
\cite{Pimkin2018} (256)                                      & 0.76                                        & \begin{tabular}[c]{@{}l@{}}ResNet34, Densenet169, Densenet201\\ XGBoost\end{tabular}                                             & \cmark                     & 12                                                                               & \begin{tabular}[c]{@{}c@{}}\pB{}\\ BreakHis\end{tabular}                                                                & 1                                                     & 300$\times$300                                                          & \xmark                                                                                                                                                                                                                                                                                                                                                                      \\\hline
\etal{Sarker} (358)                                     & 0.75                                        & Inception-v4                                                                                                                       & \cmark                     & \xmark                                                                           & \xmark                                                                                                                     & 0.083                                                 & 299$\times$299                                                          & \xmark                                                                                                                                                                                                                                                                                                                                                                      \\\hline
\cite{Rakhlin2018} (98)                                       & 0.74                                        & \begin{tabular}[c]{@{}l@{}}VGG16, ResNet-50, IcenptionV3, \\LightGBM\end{tabular}                                                 & \cmark                     & \cmark                                                                           & \xmark                                                                                                                     & \begin{tabular}[c]{@{}c@{}}0.20\\ 0.54\end{tabular}   & \begin{tabular}[c]{@{}c@{}}400$\times$400\\ 600$\times$600\end{tabular} & \begin{tabular}[c]{@{}c@{}} M3\\ \\ \end{tabular} \\\hline
\cite{Iesmantas2018} (164)                                      & 0.72                                        & \begin{tabular}[c]{@{}l@{}}Custom CNN (Capsule Network)\end{tabular}                                                             & \xmark                     & \xmark                                                                           & \xmark                                                                                                                     & 0.029                                                 & 512$\times$512                                                          & \begin{tabular}[c]{@{}c@{}} M4\\ \end{tabular}                                                                                                                                                        \\\hline
\etal{Xie} (253)                                     & 0.72                                        & CNN                                                                                                                                & \xmark                     & \xmark                                                                           & \xmark                                                                                                                     & 0.083                                                 & 512$\times$512                                                          & \xmark                                                                                                                                                                                                                                                                                                                                                                      \\\hline
\cite{Weiss2018} (268)                                      & 0.72                                        & \begin{tabular}[c]{@{}l@{}}Xception, Logistic Regression\end{tabular}                                                             & \cmark                     & \xmark                                                                           & \xmark                                                                                                                     & 1                                                     & 1024$\times$768                                                         & \begin{tabular}[c]{@{}c@{}} M3\\ \end{tabular}          \\\hline

\cite{Awan2018} (6)                                        & 0.71                                        & \begin{tabular}[c]{@{}l@{}}ResNet50, SVM\end{tabular}                                                                             & \cmark                     & \xmark                                                                           & \xmark                                                                                                                     & 0.33                                                  & 512$\times$512                                                          & \begin{tabular}[c]{@{}c@{}} M4\\ \end{tabular}                                                                                                                                                        \\\hline
Liang (62)                                      & 0.7                                         & \begin{tabular}[c]{@{}l@{}}VGG16, VGG19, ResNet50, InceptionV3,\\ Inception-Resnet, k-NN\end{tabular}                           & \cmark                     & 5                                                                                & \xmark                                                                                                                     & 0.083                                                 & N/A                                                                      & \xmark \\\hline                                                                                                                                                                                                                                                                                     
\end{tabular}
\end{table*}

%% file: table_partB.tex
\begin{table*}
\centering
\footnotesize{
\setlength{\tabcolsep}{0.4em} 
\renewcommand{\arraystretch}{1.1}
\caption{Summary of the methods submitted for whole-slide image analysis (\pB{}).
\upper{B}detailed description in Section~\ref{sec:partB};
\upper{AB}detailed description in Section~\ref{sec:partAB}. 
Pre-training is performed on ImageNet~\cite{Krizhevsky2012} unless stated otherwise. \textbf{Score} is the custom metric from Eq.~\ref{eq:score}; \textbf{Approach} lists the main methods to label the images; \textbf{Ensemble} (Ens.) indicates if the approach uses a single or multiple models (and their number, when available); \textbf{External sets} indicates if the method was trained using datasets other than from \pA{}; \textbf{Context (area ratio)} is the ratio between the average original size and the size of the patch that is used for training the network (prior to rescaling); \textbf{Input size (pixels)} is the size of the image to be analyzed by the model; \textbf{Color normalization} (Color norm.) indicates if any histology-inspired normalization was used.
$^{1}$  trained on ImageNet, $^2$ trained on VOC2012. \label{tab:partB}}

\begin{tabular}{lllcccccc}
\hline
\multicolumn{1}{c}{\textbf{Team}} & \multicolumn{1}{c}{\textbf{Score}} & \multicolumn{1}{c}{\textbf{Approach}} & \parbox{3em}{\centering\textbf{Pre- \\ trained}} & \textbf{Ens.} & \textbf{External sets} & \parbox{1.5cm}{\centering \textbf{Context \\ (area ratio)}} & \parbox{2.3cm}{\centering\textbf{Input size \\ (pixels)}} & \parbox{1cm}{\centering\textbf{Color \\ norm.}} \\ \hline

\cite{Kwok2018} (248)\upper{AB}           & 0.69         & Inception-Resnet-v2                                                                                               & \cmark                                                              & \xmark                                                          & \pA{}                                                       & 8.5e$-4$             & 299$\times$299                                                      & \xmark                                                          \\ \hline
\cite{Marami2018} (16)\upper{AB} & 0.55         & \begin{tabular}[l]{@{}l@{}}Inception-v3 + \\ adaptive pooling\end{tabular}                                     & \cmark                                                                     & 4                                                       & \begin{tabular}[c]{@{}c@{}}\pA{}\\ BreakHis\end{tabular}    & 9.9e$-5$                         & 512$\times$512                                                      & \xmark                                                          \\ \hline
Jia \textit{et al.} (296)           & 0.52         & \begin{tabular}[l]{@{}l@{}}ResNet-50 + multiscale\\  atrous convolution\end{tabular}                         & \cmark                                                              & \xmark                                                          & \xmark                                                            & 9.9e$-5$                                                           & 512$\times$512                                                       & \xmark                                                          \\ \hline
Li \textit{et al.} (137)           & 0.52         & \begin{tabular}[l]{@{}l@{}}VGG16$^1$, DeepLabV2$^1$,\\ Resnet50$^2$\end{tabular}                                  & \cmark & \cmark                                                 & \xmark                                                            & 9.9e$-5$                                                            & 512$\times$512                                                       & \xmark                                                          \\ \hline
Murata \textit{et al.} (91)            & 0.50         & U-Net                                                                                                             & \xmark                                                                & \xmark                                                          & \xmark                                                            & 1.6e$-2$ & 256$\times$256                                                       & \xmark                                                          \\ \hline
\cite{Galal2018} (264)\upper{B}           & 0.50         & DenseNet                                                                                                          & \xmark                                                             & \xmark                                                          & \xmark                                                            & 1.6e$-3$                           & 2048$\times$2048                                                     & \xmark                                                          \\ \hline
\cite{Vu2018} (166)\upper{AB}  & 0.49         & \begin{tabular}[l]{@{}l@{}} DenseNet, SENet, ResNext\end{tabular} & \cmark                                                              & \xmark                                                          & \xmark                                                            & 1.5e$-4$                                                                                   & $630\times 630$                                                      & \xmark                                                          \\  \hline
\cite{Kohl2018} (54)\upper{AB}                             & 0.42                                        & Densenet-161  & \cmark                                                              & \xmark                                                          & \begin{tabular}[c]{@{}c@{}}\pB{} non- \\annotated \end{tabular}                                                            & 9.3e$-6$                                                                                   & $157\times 157$                                                      & \xmark                                                          \\  \hline

\end{tabular}
}
\end{table*}

%% file: results.tex
\section{Results}
\label{sec:results}

\begin{figure*}[t]
\centering
\begin{subfigure}{0.48\textwidth}
\includegraphics[width=\textwidth]{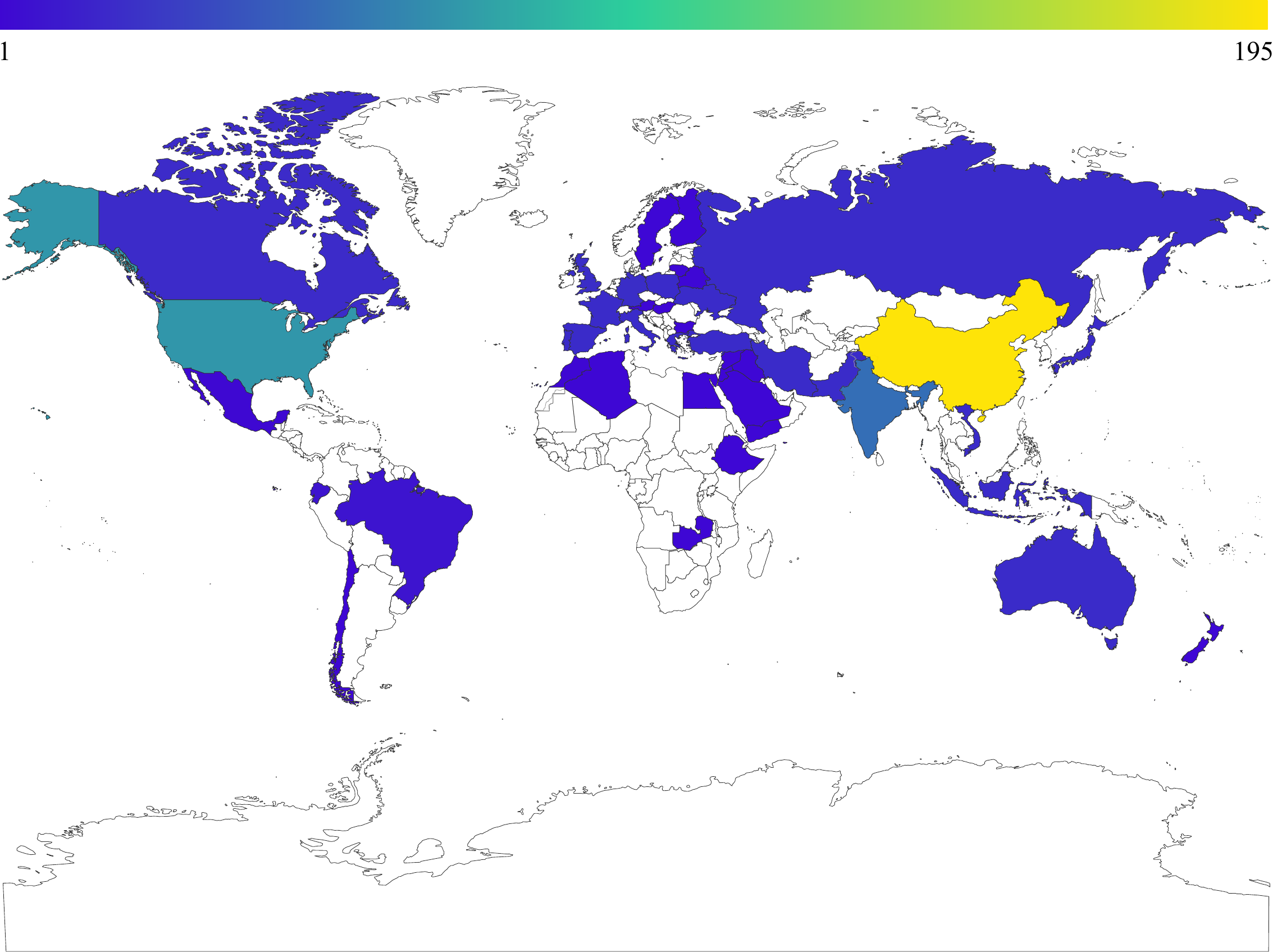}
\caption{Registrations.}
\end{subfigure}
\hfill
\begin{subfigure}{0.48\textwidth}
\includegraphics[width=\textwidth]{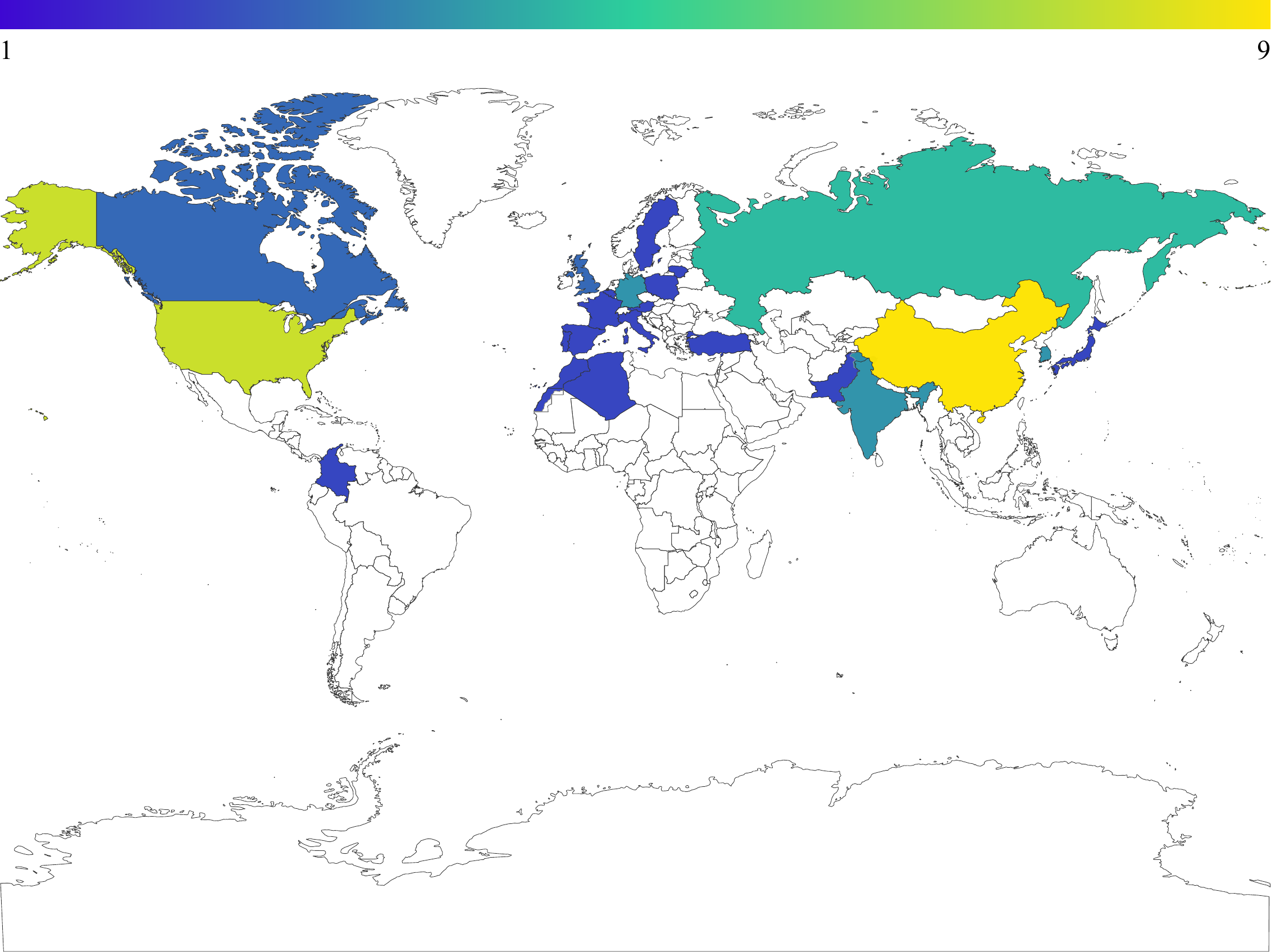}
\caption{Submissions.}
\end{subfigure}
\hfill
\caption{Geographical distribution of the BACH participants. a) registered on the website; b) submitted for the test set.  \label{fig:participants}}
\end{figure*}

The BACH had worldwide participation, with a total of 677 registrations and 64 submissions for both part A (51) and B (13), as shown on Fig.~\ref{fig:participants}.

\subsection{Performance in \pA{}}

Participants of the \pA{} of the challenge were ranked in terms of accuracy. As in \etal{Ara\'{u}jo}~\cite{Araujo2017}, these submissions were further evaluated in terms of sensitivity and specificity:

\begin{equation}
\text{sensitivity} = \frac{TP}{TP+FN}
\end{equation}

\begin{equation}
\text{specificity} = \frac{TN}{TN+FP}
\end{equation}

\noindent where $TP$, $TN$, $FP$ and $FN$ are the class-wise true-positive, true-negative, false-positive and false-negative predictions, respectively. For benchmarking purposes, a simple fine-tuning experiment on the BACH \pA{} was conducted. Specifically, the classification parts of VGG16, Inception~v3, ResNet50 and DenseNet169 were replaced by a pair of untrained fully-connected layers with 1024 and 4 neurons. These networks were then trained on two steps, first by updating only the new fully-connected layers until the validation loss stopped improving and posteriorly training the entire model until the same stop criteria was met. Adam~\cite{Kingma2015} was used as optimizer and the loss was the categorical cross-entropy.
\par
Finally, to further evaluate the performance of the methods submitted to \pA{}, four pathologists (E1--E4) were asked to classify the field images from the BACH test set. E1 and E3 are breast cancer specialists and E2 and E4 are experienced pathologists. Also, E1 was one of the two experts involved in the construction of training and testing sets from BACH, and the remaining three were external to the process. The difference in the annotation process was that in the BACH sets construction the pathologists had access to other regions of the patient tissue (and potentially imunohistochemical analysis), whether in this second phase they could only see the field image, \textit{i.e.}, they only had access to the same information of the automated classification algorithms. 
\par
The class-wise performance of the methods is shown in Table~\ref{tab:pA_sens}. Table~\ref{tab:pA_sens_2} shows the performance for two binary cases:
\begin{inparaenum}[1)]
\item a referral scenario, \textit{Pathological}, where the objective is to distinguish Normal images from the remaining classes and,
\item a cancer detection scenario, \textit{Cancer}, where the Normal and Benign classes are grouped \textit{vs} the \textit{In situ} and Invasive classes.
\end{inparaenum}
\par

\begin{table}[t]
\centering
\caption{Class-wise sensitivity and specificity of \pA{} approaches for the classes in study. Benchmarking results via fine-tuning are also shown. Acc - accuracy; Se - sensitivity; Sp - specificity. Expert 1 annotated the BACH dataset. \label{tab:pA_sens}}
\setlength{\tabcolsep}{0.15em} 
\footnotesize
\begin{tabular}{c|c||c|c|c|c|c|c|c|c}\hline

\multicolumn{2}{c|}{}& \multicolumn{2}{|c}{\textbf{Normal}}& \multicolumn{2}{c}{\textbf{Benign}} & \multicolumn{2}{c}{\textbf{\textit{In situ}}}  & \multicolumn{2}{c}{\textbf{Invasive}}\\\hline
\textbf{Team} & \textbf{Acc}  & Se. & Sp. & Se. & Sp.& Se. & Sp.& Se. & Sp.\\\hline
\textbf{216} &0.87 & 0.96& 0.88& 0.8& 0.96& 0.84& 1.0& 0.88& 0.99\\
\textbf{248} &0.87 & 0.96& 0.93& 0.72& 0.96& 0.88& 0.97& 0.92& 0.96\\
\textbf{1} &0.86 & 0.96& 0.91& 0.68& 0.97& 0.84& 0.99& 0.96& 0.95\\
\textbf{16} &0.84 & 0.92& 0.95& 0.64& 0.96& 0.84& 0.99& 0.96& 0.89\\
\textbf{54} &0.83 & 0.96& 0.92& 0.52& 0.97& 0.88& 0.92& 0.96& 0.96\\
\textbf{157} &0.83 & 0.96& 0.91& 0.64& 0.99& 0.92& 0.91& 0.8& 0.97\\
\textbf{186} &0.81 & 0.96& 0.92& 0.68& 0.96& 0.76& 0.95& 0.84& 0.92\\
\textbf{19} &0.81 & 1.0& 0.95& 0.4& 0.99& 0.92& 0.92& 0.92& 0.89\\
\textbf{36} &0.81 & 0.88& 0.92& 0.6& 0.96& 0.88& 0.95& 0.88& 0.92\\
\textbf{412} &0.8 & 0.92& 0.96& 0.48& 0.97& 0.84& 0.92& 0.96& 0.88\\\hline
VGG & 0.58 & 0.84  & 0.84 & 0.64 & 0.84 & 0.72 & 0.87 & 0.36 & 0.97 \\
Inception & 0.77 & 0.92 & 0.93 & 0.44 & 0.96 & 0.88 & 0.87 & 0.84 & 0.93 \\
ResNet & 0.76 & 0.88 & 0.92 & 0.52 & 0.95 & 0.8 & 0.87 & 0.84 & 0.95 \\
DenseNet & 0.79 & 0.92 & 0.96 & 0.36 & 0.99 & 0.92 & 0.83 & 0.96 & 0.95 \\ \hline
Expert 1 & 0.96 & 0.96 & 0.99 & 0.92 & 0.97 & 1.0 & 1.0 & 0.96 & 0.99 \\
Expert 2 & 0.94 & 0.96 & 0.99 & 0.88 & 0.96 & 1.0 & 1.0 & 0.92 & 0.97 \\
Expert 3 & 0.78 & 0.88 & 0.99 & 0.76 & 0.79 & 0.56 & 0.97 & 0.92 & 0.96 \\
Expert 4 & 0.73 & 0.40 & 0.99 & 0.84 & 0.71 & 0.76 & 0.97 & 0.92 & 0.97 \\ \hdashline
\parbox{5em}{\centering\textbf{Experts}\\\textbf{(avg.)}} & \splito{0.85$\pm$}{0.10} & \splito{0.80$\pm$}{0.23} & \splito{0.99$\pm$}{0.00} & \splito{0.85$\pm$}{0.06} & \splito{0.86$\pm$}{ 0.11} & \splito{0.83$\pm$}{0.18} & \splito{0.99$\pm$}{0.01} & \splito{0.93$\pm$}{0.02} & \splito{0.97$\pm$}{0.01} \\\hline
\end{tabular}
\end{table}

\begin{table}[t]
\centering
\caption{Class-wise sensitivity and specificity of \pA{} approaches. \textbf{Patho.} refers to Benign, \textit{in situ} and Invasive \textit{vs} Normal, and \textbf{Cancer} refers to \textit{in situ} and Invasive \textit{vs} Normal and Benign classes. Benchmarking results via fine-tuning are also shown. Acc - accuracy ($\pm$ the confidence interval); Se - sensitivity; Sp - specificity. Expert 1 annotated the BACH dataset. \label{tab:pA_sens_2}}
\setlength{\tabcolsep}{0.35em}
\footnotesize
\begin{tabular}{c|c|c|c||c|c|c}\hline

\multicolumn{1}{c}{}&\multicolumn{3}{c}{\textbf{Patho.}} & \multicolumn{3}{c}{\textbf{Cancer}} \\\hline
\textbf{Team} & Acc. & Se. & Sp. & Acc.& Se. & Sp.\\\hline
\textbf{216} & 0.9& 0.88& 0.96& 0.92 & 0.86& 0.98\\
\textbf{248} &0.94 &  0.93& 0.96& 0.92 & 0.92& 0.92\\
\textbf{1}  & 0.92 & 0.91& 0.96& 0.9 & 0.9& 0.9\\
\textbf{16} & 0.94 & 0.95& 0.92&0.9 &0.94& 0.86\\
\textbf{54} & 0.93 & 0.92& 0.96& 0.89 & 0.94& 0.84\\
\textbf{157} & 0.92 & 0.91& 0.96& 0.94 & 0.96& 0.92\\
\textbf{186} & 0.93 &  0.92& 0.96& 0.9 & 0.9& 0.9\\
\textbf{19}  & 0.96 & 0.95& 1.0& 0.86 & 0.96& 0.76\\
\textbf{36} & 0.91 & 0.92& 0.88& 86 & 0.9& 0.82\\
\textbf{412} & 0.95 & 0.96& 0.92& 0.86 & 0.96& 0.76\\\hline
VGG16 & 0.84& 0.84 & 0.84 &0.66  & 0.66  & 0.88 \\
Inception V3 & 0.93 & 0.93 & 0.92 &0.92 & 0.92 & 0.76 \\
ResNet 50 & 0.92 & 0.92  & 0.88 & 0.86  & 0.86  & 0.76  \\
DenseNet 169 & 0.96 & 0.96 & 0.92 & 0.96 & 0.96 & 0.68 \\ \hline
Expert 1 & 0.98 & 0.99 & 0.96 & 0.98 & 0.98 & 0.98 \\
Expert 2 & 0.98 & 0.99 & 0.96 & 0.96 & 0.96 & 0.96 \\
Expert 3 & 0.96 & 0.99 & 0.88 & 0.82 & 0.74 & 0.90 \\
Expert 4 & 0.84 & 0.99 & 0.40 & 0.90 & 0.86 & 0.94 \\\hdashline
\parbox{5em}{\centering\textbf{Experts}\\\textbf{(avg.)}} & \splito{0.94$\pm$}{0.06} & \splito{0.99$\pm$}{0.00} & \splito{0.80$\pm$}{0.24} & \splito{0.91$\pm$}{0.06} & \splito{0.88$\pm$}{0.09} & \splito{0.95$\pm$}{0.03}\\\hline
\end{tabular}
\end{table}

\begin{figure*}[htpb!]
\centering

\begin{subfigure}{0.4\textwidth}
\centering
\includegraphics[width=\textwidth]{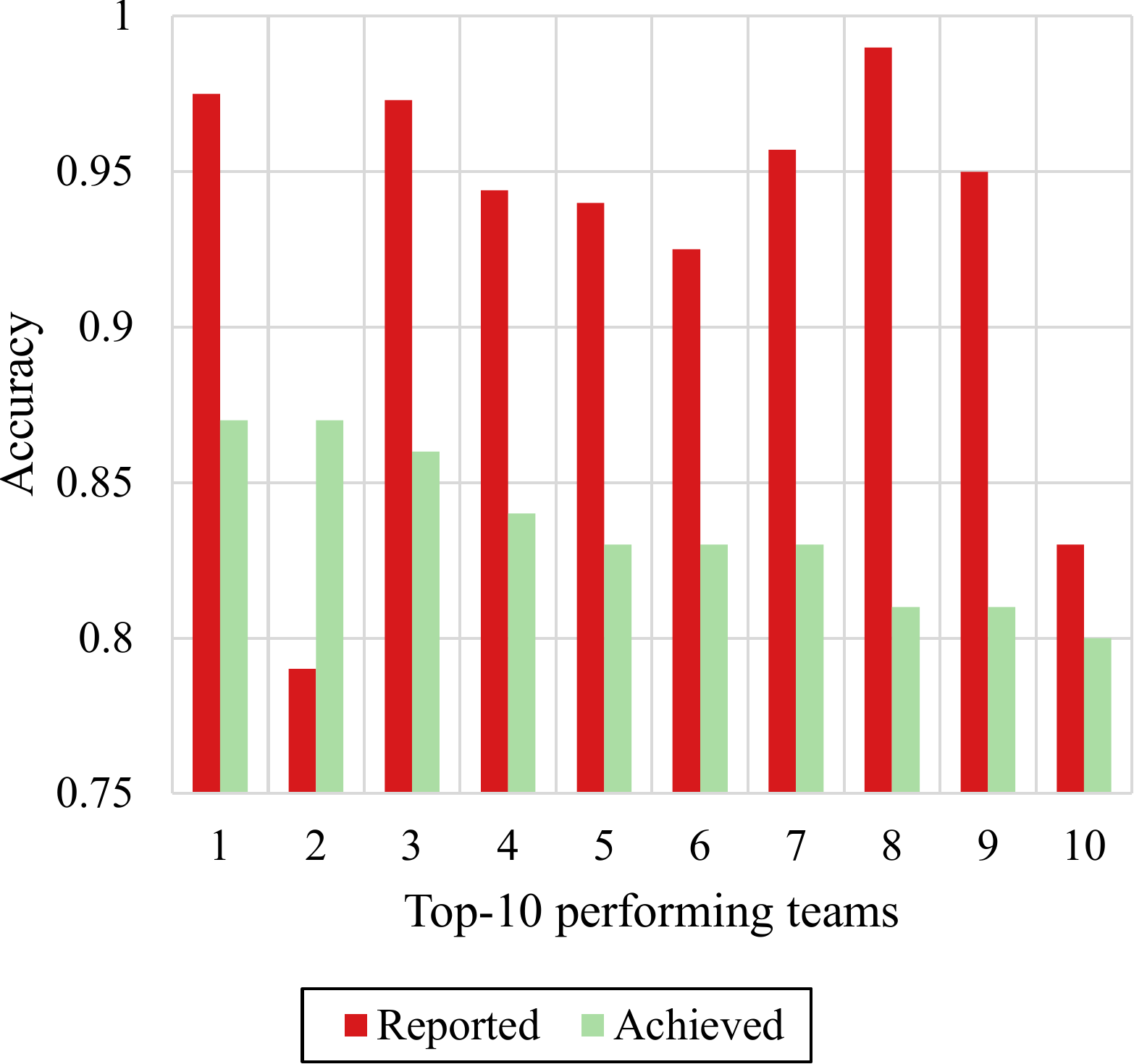}
\caption{Reported performance during the method submission and respective test accuracy for the top-10 performing teams. \label{fig:reported_performance}}
\end{subfigure}
\qquad
\begin{subfigure}{0.4\textwidth}
\includegraphics[width=\textwidth]{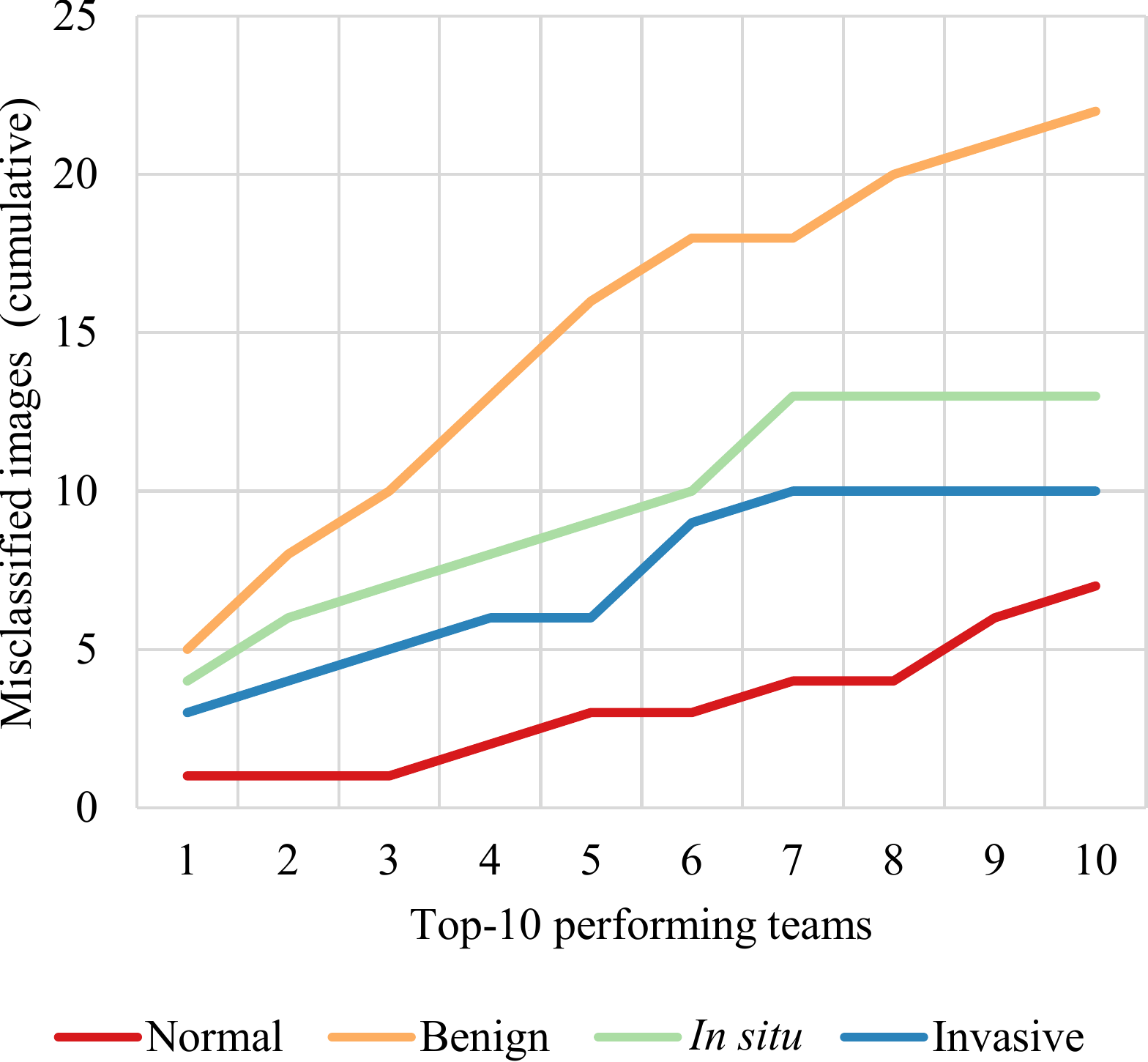}
\caption{Cumulative number of unique misclassifications per class for the top-10 performing teams. Higher values are indicative of a more challenging classification.\label{fig:misclassifications}} 
\end{subfigure}

\begin{subfigure}{0.4\textwidth}
\includegraphics[width=\textwidth]{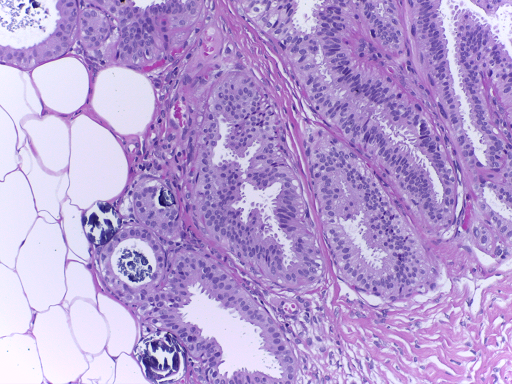}
\caption{Case of Benign mostly misclassified as \textit{In situ}\label{fig:most_erroneous1}}
\end{subfigure}
\qquad
\begin{subfigure}{0.4\textwidth}
\includegraphics[width=\textwidth]{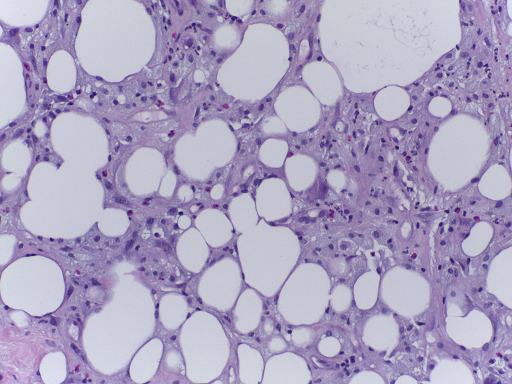}
\caption{Case of Benign mostly misclassified as Invasive\label{fig:most_erroneous2}}
\end{subfigure}

\begin{subfigure}{0.4\textwidth}
\includegraphics[width=\textwidth]{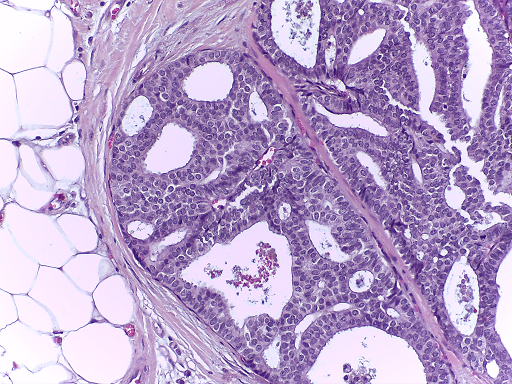}
\caption{Example of an \textit{In situ} case from the training set\label{fig:most_erroneous1_train}}
\end{subfigure}
\qquad
\begin{subfigure}{0.4\textwidth}
\includegraphics[width=\textwidth]{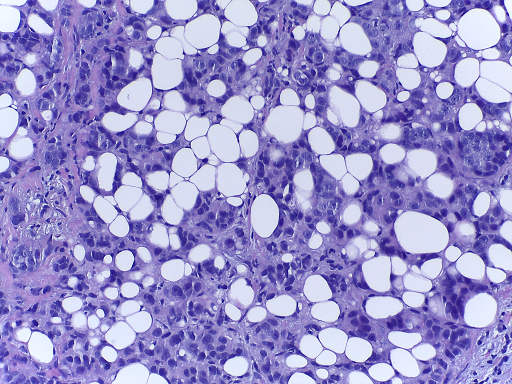}
\caption{Example of an Invasive case from the training set\label{fig:most_erroneous2_train}}
\end{subfigure}

\caption{Examples of images misclassified by the top-10 methods of \pA{} and similar examples in the training set.\label{fig:most_erroneous}}
\end{figure*}

Fig.~\ref{fig:reported_performance} depicts, for the top-10 participants, the difference between the reported performances on the training set (cross-validation) and the achieved performances on the hidden test set. Also, a class-wise study of these methods shows that the Benign and \textit{In situ} classes are the most challenging to classify (Fig.~\ref{fig:misclassifications}). In particular, Fig.~\ref{fig:most_erroneous1}~and~\ref{fig:most_erroneous2} show two images with 100\% inter-observer agreement that were misclassified by the majority (at least 80\%) of the top-10 approaches. 

\subsubsection{Inter-observer analysis}

\begin{figure*}[htpb!]
    \centering
    \includegraphics[width=\textwidth]{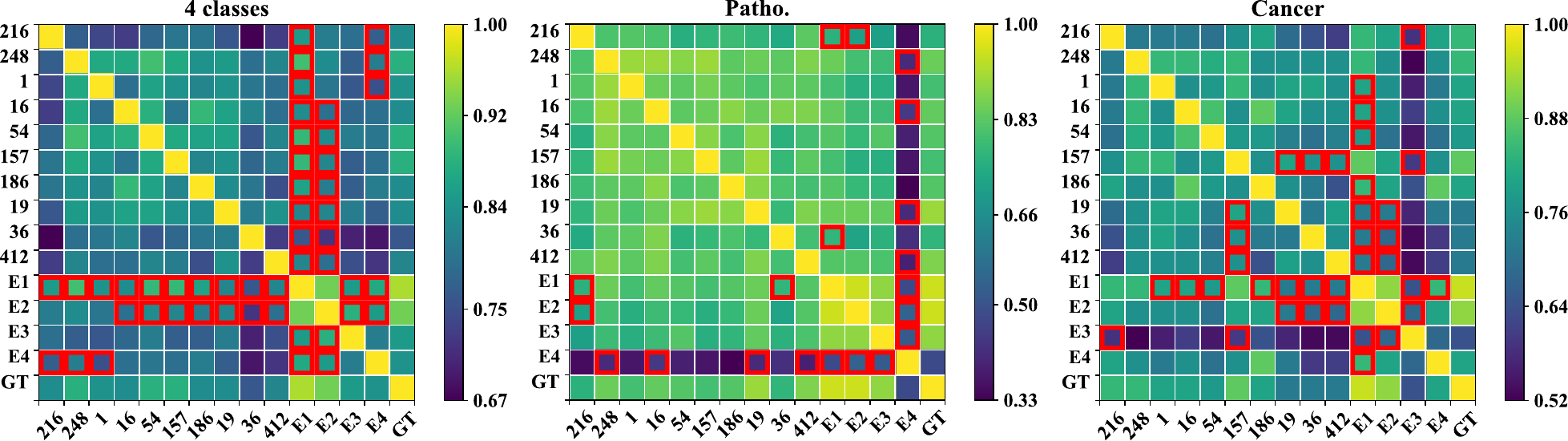}
    \caption{Inter-observer, inter-method and observer-method quadratic-weighted kappa score \pA{}. E1--E4 are four expert pathologists, GT is the ground-truth and the numbers indicate the respective team. Expert 1 annotated the BACH dataset one month before participating in the inter-observer study. \textbf{4 classes} considers the 4 classes used in this study. \textbf{Patho.} refers to Benign, \textit{In situ} and Invasive \textit{vs} Normal, and \textbf{Cancer} refers to \textit{In situ} and Invasive \textit{vs} Normal and Benign classes. {\color{red}$\Box$}$~$the prediction is statistically different according to the McNemar's test.}
    \label{fig:kappa_pA}
\end{figure*}

\begin{figure}
    \centering
    \includegraphics[width=\columnwidth]{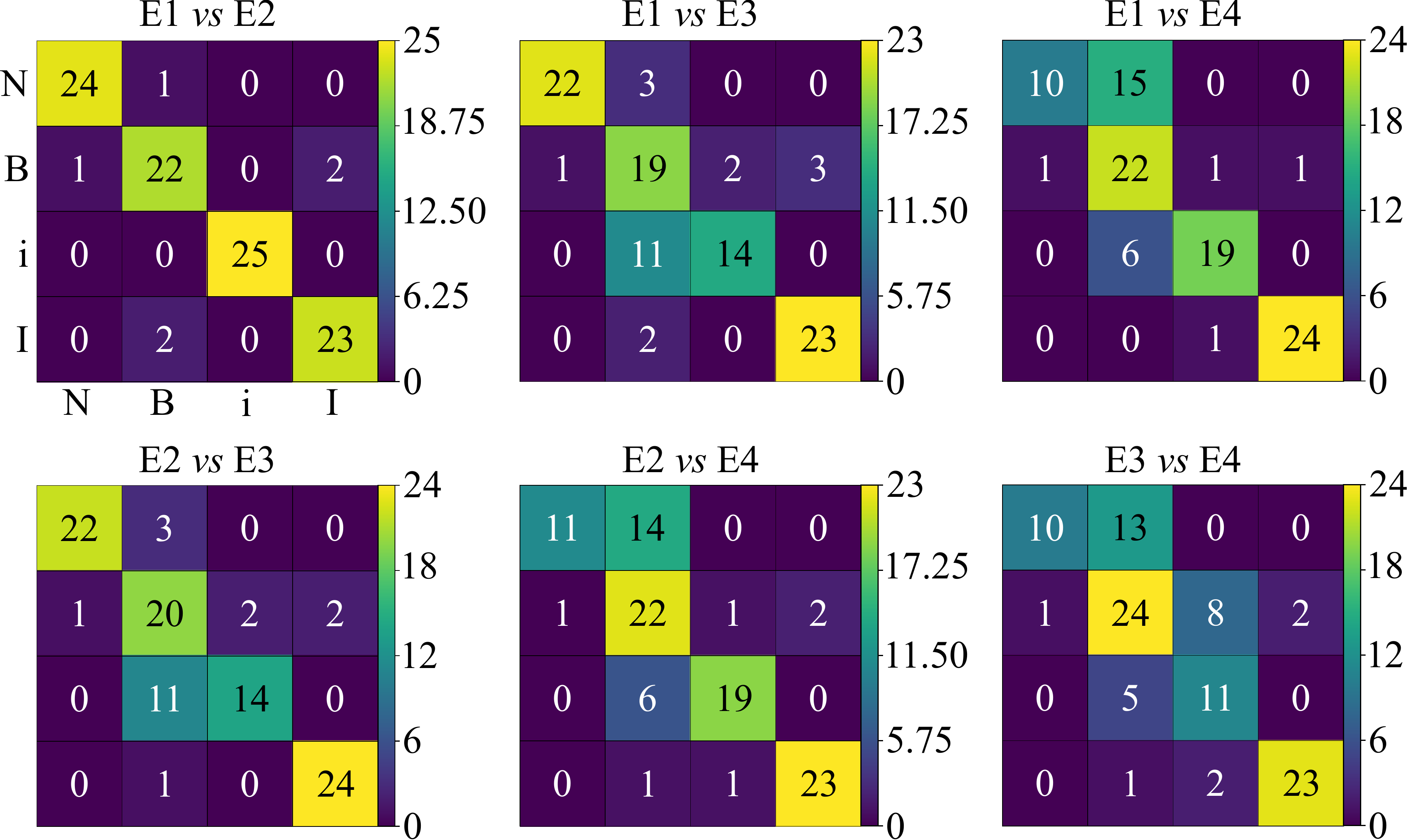}
    \caption{Confusion matrices of the 4 expert annotators (E\#) for the 100 test images of \pA{}. \textbf{N} - Normal; \textbf{B} - Benign; \textbf{i} - \textit{in situ}; \textbf{I} - Invasive.}
    \label{fig:conf_A}
\end{figure}

The accuracies of the three external pathologists are of 94$\%$, 78$\%$ and 73$\%$, and the accuracy of the pathologist from BACH is 96$\%$. Note that this pathologist annotated the images after one month from the first annotation, in order to avoid the influence of past knowledge regarding the patients' exams. For comparison purposes, Fig.~\ref{fig:kappa_pA} shows the inter-observer, inter-method and observer-method agreement via the quadratic-weighted Cohen's kappa score and their corresponding statistical differences, and Fig.~\ref{fig:conf_A} shows the confusion matrices of the experts.

\subsection{Performance in \pB{}}

\begin{figure*}[tb]
\centering
\begin{subfigure}{0.45\textwidth}
\includegraphics[width=\textwidth]{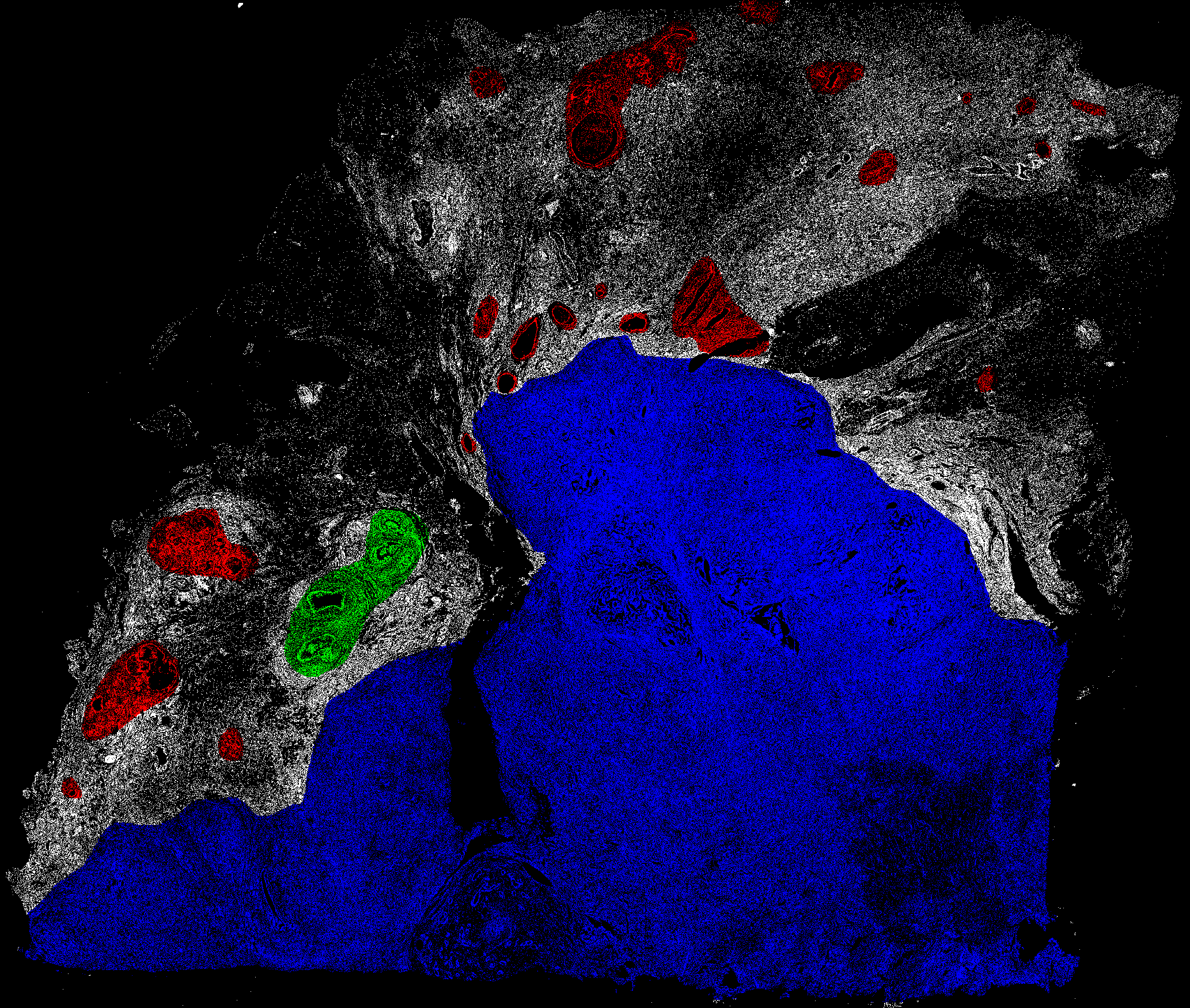}
\caption{Ground-truth (image 10). }
\end{subfigure}
\hfill
\begin{subfigure}{0.45\textwidth}
\includegraphics[width=\textwidth]{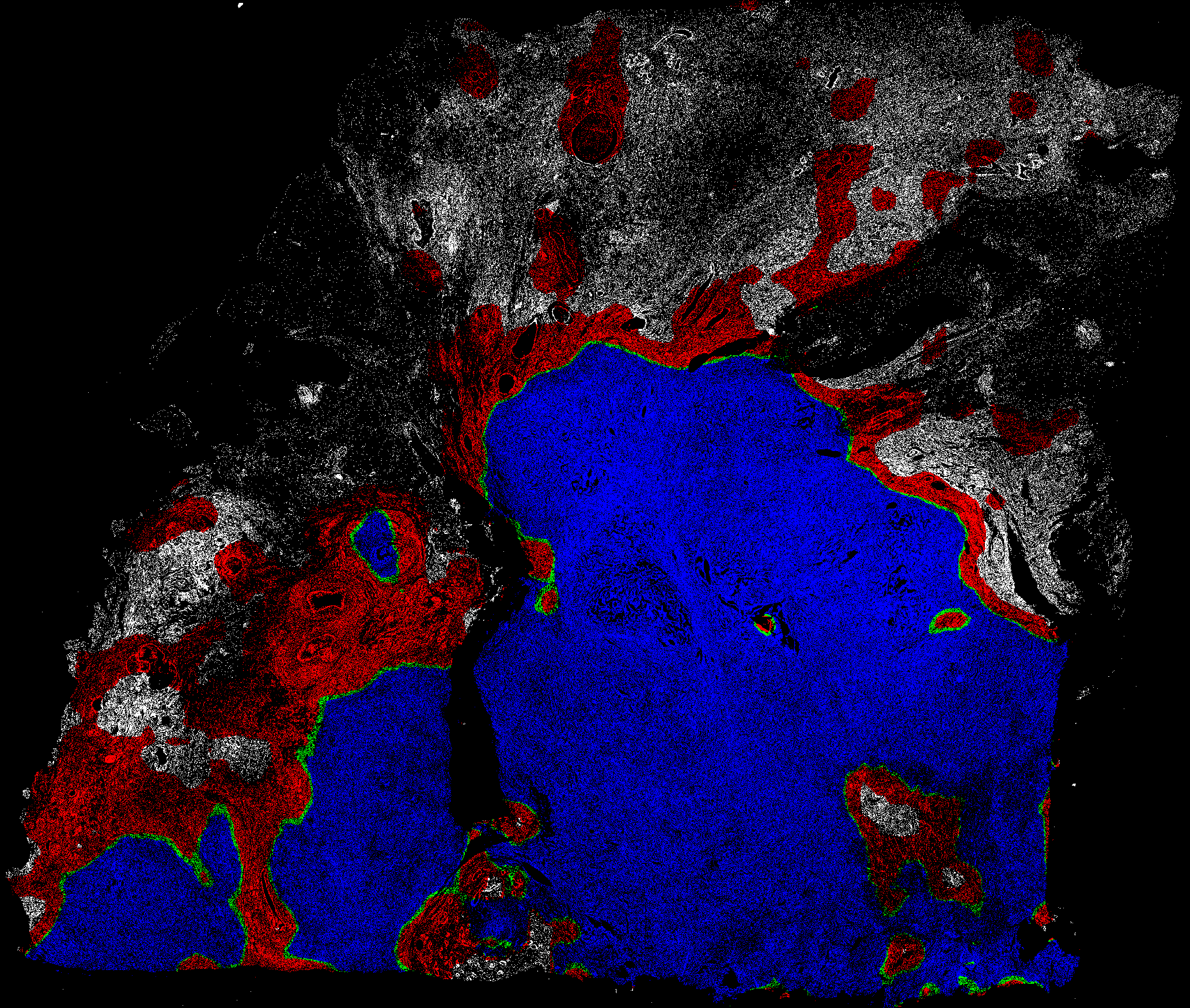}
\caption{Prediction from team 248. $s=0.897$}
\end{subfigure}

\begin{subfigure}{0.45\textwidth}
\includegraphics[width=\textwidth]{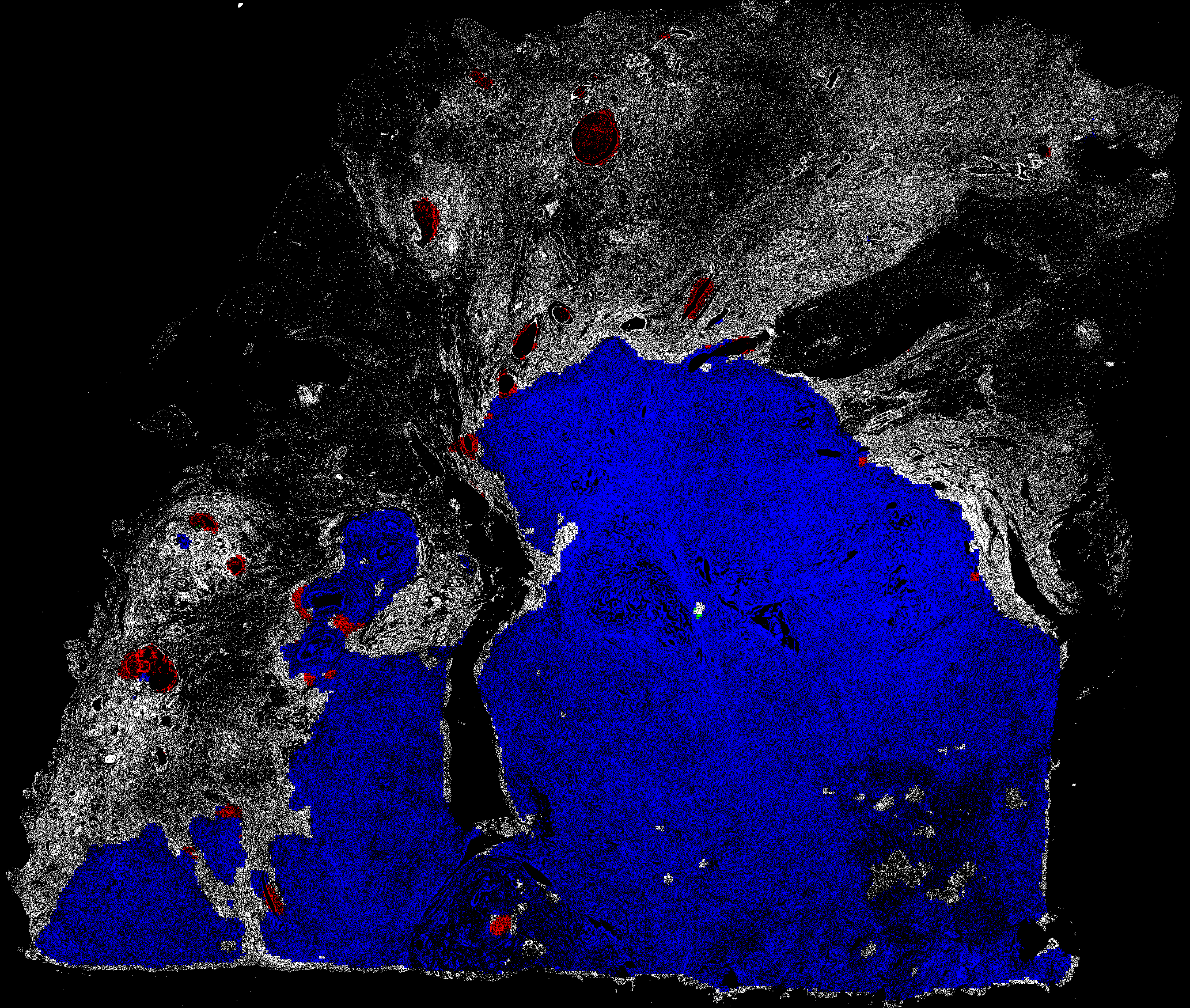}
\caption{Prediction from team 16. $s=0.842$}
\end{subfigure}
\hfill
\begin{subfigure}{0.45\textwidth}
\includegraphics[width=\textwidth]{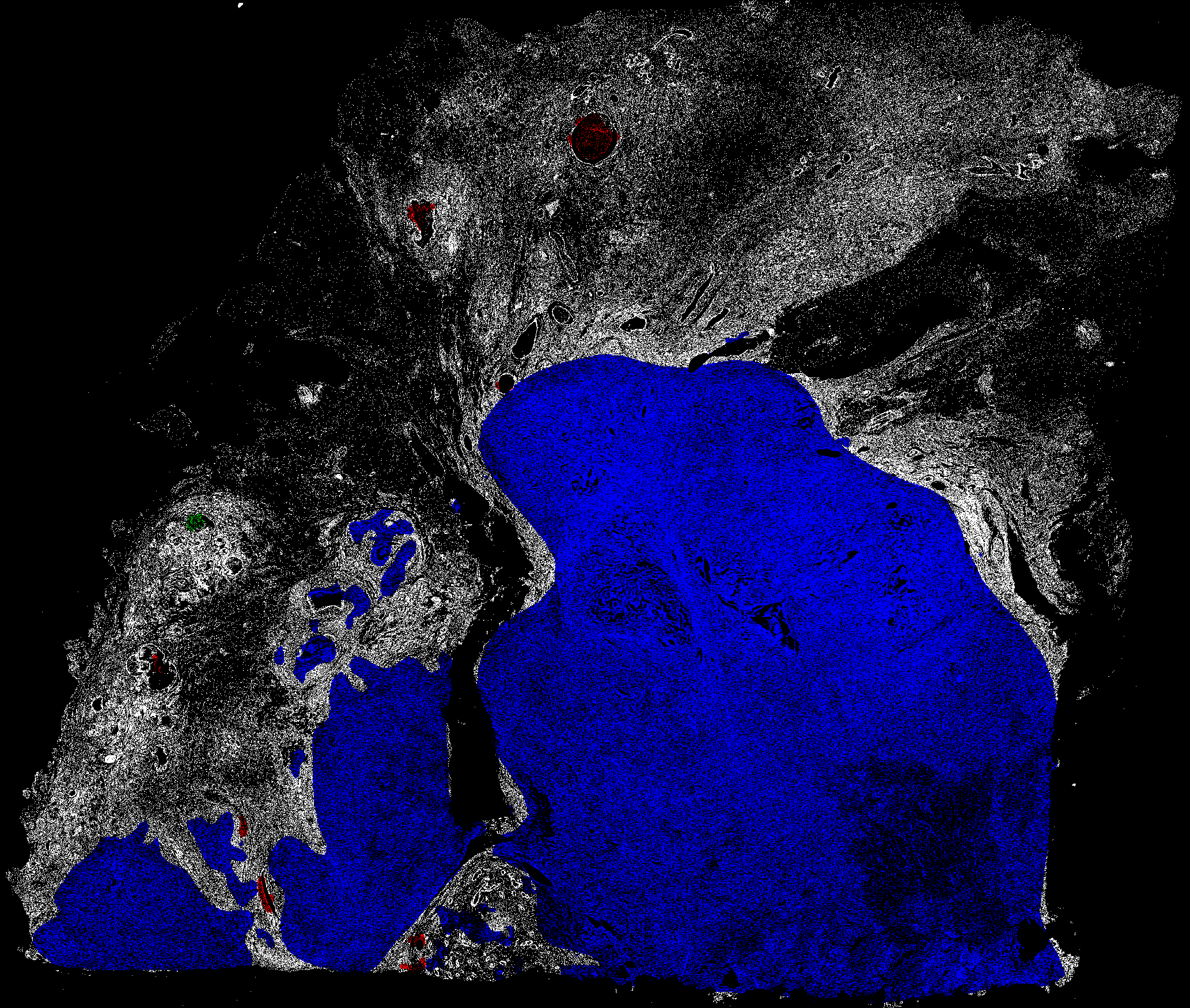}
\caption{Prediction from team 296. $s=0.905$}
\end{subfigure}

\caption{Examples of test set predictions for \pB{} from the top performing teams. {\color{red}$\blacksquare$}$~$benign; {\color{green}$\blacksquare$}$~$\textit{in situ}; {\color{blue}$\blacksquare$}$~$invasive. The original WSIs were converted to grayscale and the teams' predictions overlayed (background was removed, appearing in black). The obtained $s$ scores (Eq.~\ref{eq:score}) are also shown. \label{fig:pb_pred}}
\end{figure*}

The overall challenge performance and main approaches of the participating teams are shown in Table~\ref{tab:partB}. Similarly to \pA{}, please refer to Tables~\ref{tab:pB_sens}~and~\ref{tab:pB_sens_2} for the team-wise sensitivity and specificity of the methods. For reference purposes, Table~\ref{tab:pB_sens} also shows the quadratic-weighted kappa scores for each method.
Examples of pixel-wise predictions for \pB{} are shown in Fig.~\ref{fig:pb_pred}. The correct identification of invasive regions was more successful, as opposed to benign and 	\textit{in situ} regions.

\begin{table}[tb]
\centering
\caption{Class-wise sensitivity and specificity of \pB{} approaches for the classes in study, along with the challenge score and the quadratic-weighted kappa score ($\pm$ the confidence interval). Se - sensitivity; Sp - specificity; Score - challenge score $s$ (Eq.~\ref{eq:score}); $k$ - kappa score. \label{tab:pB_sens}}
\footnotesize

\begin{tabular}{c|c|c||c|c|c|c|c|c|c}\hline

\multicolumn{3}{c|}{}& \multicolumn{2}{|c}{\textbf{Benign}}& \multicolumn{2}{c}{\textbf{\textit{In situ}}} & \multicolumn{2}{c}{\textbf{Invasive}}  \\\hline
\textbf{Team} & \textbf{Score} & $k$ & Se. & Sp. & Se. & Sp.& Se. & Sp.\\\hline
\textbf{248}& 0.69$\pm0.07$ &0.44$\pm0.18$	&0.36&	0.7&	0.03&	0.59&	0.4&	0.96\\
\textbf{16} & 0.55$\pm0.11$ &0.51$\pm0.15$	&0.09&	0.99&	0.05&	0.95&	0.45&	0.92	\\
\textbf{296} & 0.52$\pm0.16$&0.48$\pm0.19$ & 0.07& 0.99&	0.03&	0.95&	0.39&	0.93\\
\textbf{137} & 0.52$\pm0.15$&0.51$\pm0.18$ & 0.04& 1&	0.02&	0.92&	0.53&	0.89\\
\textbf{91}	 & 0.50$\pm0.10$&0.28$\pm0.12$ & 0.05& 0.8&	0.18&	0.53&	0.13&	0.89\\
\textbf{264}& 0.50$\pm0.13$ &0.29$\pm0.17$ &0.05&	0.88&	0.08&	0.52&	0.47&	0.74   \\
\textbf{166}& 0.49$\pm0.11$&0.28$\pm0.15$ & 0.14&	0.9&	0.05&	0.63&	0.44&	0.76\\
\textbf{94}&  0.47$\pm0.15$&0.39$\pm0.18$ & 0.16&	0.95&	0.05&	0.76&	0.5&	0.78\\
\textbf{256}& 0.46$\pm0.14$&0.17$\pm0.14$ & 0.18&	0.58&	0&	    0	&   0.58&	0.47\\
\textbf{15}&  0.46$\pm0.13$&0.27$\pm0.15$ & 0.11&	0.78&	0.02&	0.46&	0.4&	0.68\\
\textbf{54} &  0.42$\pm0.14$&0.34$\pm0.15$ & 0.03&	0.98&	0.06&	0.75&	0.52&	0.74\\
\textbf{183}  & 0.39$\pm0.10$&0.12$\pm0.13$ & 0.31&	0.81&	0.15&	0.47&	0.15&	0.71\\
\textbf{252}& 0.33$\pm0.12$&0.13$\pm0.09$ & 0.02&	0.98&	0&	    0.85&	0.2&	0.88\\ \hline

\end{tabular}

\end{table}

\begin{table}[tb]
\centering
\caption{Class-wise sensitivity and specificity of \pB{} approaches. \textbf{Patho.} refers to Benign, \textit{In situ} and Invasive \textit{vs} Normal, and \textbf{Cancer} refers to \textit{In situ} and Invasive \textit{vs} Normal and Benign classes. Se - sensitivity; Sp - specificity. \label{tab:pB_sens_2}}
\footnotesize
\begin{tabular}{c|c|c|c|c}\hline

\multicolumn{1}{c}{} & \multicolumn{2}{|c|}{\textbf{Patho.}} & \multicolumn{2}{c}{\textbf{Cancer}} \\\hline
\textbf{Team}  & Se. & Sp.& Se. & Sp. \\\hline
\textbf{248}&	0.78&	0.59&	0.46&	0.93\\
\textbf{16} &	0.6&	0.95&	0.52&	0.93\\
\textbf{296}& 0.53&	0.95&	0.43&	0.93\\
\textbf{137}& 0.63&	0.92&	0.58&	0.89\\
\textbf{91}&0.71&	0.53&	0.55&	0.71\\
\textbf{264} &0.81&	0.52&	0.61&	0.58\\
\textbf{166} &	0.68&	0.63&	0.52&	0.72\\
\textbf{94}& 0.7&	0.76&	0.58&	0.78\\
\textbf{256} &	0.9&	0   &   0.68&	0.48\\
\textbf{15}& 0.82&	0.46&	0.56&	0.65\\
\textbf{54}  & 0.68&	0.75&	0.57&	0.74\\
\textbf{183} & 0.8&	0.47&	0.45&	0.61\\
\textbf{252}& 0.3&	0.85&	0.29&	0.87\\ \hline
\end{tabular}

\end{table}

\subsection{Statistical analysis}

The top-10 methods of \pA{} and the annotations of the medical experts are statistically compared via an adaption of McNemar's test~\cite{Edwards1948, Dietterich1993}. The McNemar test is based on the chi-squared distribution and allows to assess the performance of two classifiers based on their accuracy on an independent test set. Let A and B be two methods to compare. The chi-squared ($\chi^2$) distribution with 1 degree of freedom is defined by Eq.~\ref{eq:mcnemar}:

\begin{equation}
   \chi^2 = \frac{(\,|\,n_{01}-n_{10}\,|-1\,)^2}{n_{01}+n_{10}}
   \label{eq:mcnemar}
\end{equation}

\noindent where $n_{01}$ is the number of misclassified test samples by B but not by A and $n_{10}$ is the number of samples misclassified by A but not by B.
The null assumption that A and B have equal classification performance is rejected if $\chi^2>3.841$, corresponding to a p-value of 0.05. The statistical analysis for \pA{} is summarized in Fig.~\ref{fig:kappa_pA}.

The \pB's submissions are statistically assessed by the confidence intervals of their average performance in terms of the BACH score $s$ and quadratic-weighted kappa score. Namely, assuming that the scores belong to a Gaussian distribution, the confidence interval $ci$ is computed as in Eq.~\ref{eq:conf_inter}:

\begin{equation}
    ci_m = \pm 1.96\frac{\sigma_m}{\sqrt{n}}
    \label{eq:conf_inter}
\end{equation}

\noindent where 1.96 is the critical value of the confidence interval for $p=0.05$, $\sigma_m$ is the standard deviation of one of the studied scores for method $m$ and $n$ is the size of the population (10 for this study).

%% file: discussion.tex
\section{Discussion}

BACH accounted for a large number of final submissions in comparison to other medical image challenges. Despite this, there is a similar significant difference between the number of registrations and effective submissions. This stems from common factors such as 
\begin{inparaenum}[1)]
\item{registrations to inspect the data before deciding to participate or to get the data for other purposes,}
\item{difficulty in downloading the data, which is specially true in countries with Internet accessibility limitations}, and
\item{high complexity of the task, specially of part B.}
\end{inparaenum}
The verified drop on the submission rate is common on biomedical imaging challenges \footnote{\url{https://grand-challenge.org/challenges}}, which points to a need to revise future challenge designs to keep the participants' interest throughout its entire duration.
BACH, similarly to other medical image challenges, partially addressed this issue by partnering with the ICIAR conference, which empirically motivated participants by providing an opportunity to show their developed work to the scientific community. For future challenge organizations, it will be needed to further promote participation not only by improving data access but also, for instance, establishing intermediary benchmark timepoints in which participants can compare their performance to motivate themselves to further improve their methods.

The vast majority of the submitted methods used deep learning for solving both tasks A and B. This follows the common trend on the field of medical image analysis, where deep learning approaches are complementing or even replacing the standard manual feature engineering approaches since they allow to achieve high performances while significantly reducing the need for field-knowledge~\cite{Litjens2017}.
As known, deep learning approaches require large amounts of training data to produce a generalizable model, which are usually not available for medical image analysis due to complexity and high cost of the annotation process. As a consequence, it is a common practice to initialize the models with filters trained on large datasets of natural images, such as ImageNet, and fine-tune them to the specific problem \cite{Tajbakhsh2016}. In fact, as shown in Table~\ref{tab:partA}, all of the top performing methods are composed by one or more deep CNNs architectures such as Inception~\cite{Szegedy2015}, DenseNet~\cite{Huang2017}, VGG~\cite{Simonyan2014} or ResNet~\cite{He2016} pre-trained on ImageNet. 
The difference in performance is thus mainly a consequence of design and training details. 
For \pA{}, and unlike previous approaches for the analysis of breast histology cancer images \cite{Araujo2017}, the results of BACH suggest that training the models with a large portion/entire image (even if resized to fit the standard input size of the network) is better than using local patches. This indicates that the overall nuclei and tissue organization may be more relevant than nuclei-scale features, such as nuclei texture, for distinguishing different types of breast cancer. Interestingly this matches the importance that clinical pathologists give to tissue architecture features in the diagnostic task. In fact, unlike small patch-based approaches, using large portions of the images eases the integration of both local and global context in the decision process. 
Besides, patch-based approaches have to handle the problem of patch-label attribution based on the image-level label. Although more sophisticated methods, such as Multiple Instance Learning-based ones could be applied, the vast majority of the teams attributes the label of the image to the patch, which has obvious limitations since the patch may contain only normal tissue, for instance, and be labeled with a different class. 
\par
For \pB{}, the large image size inhibits the direct application of standard segmentation networks, such as U-Net \cite{Ronneberger2015}, to the entire image. Consequently, participants dealt with the issue by analyzing local patches and performing a posterior fusion of the outputs to produce the final probability mapping. In fact, following the same trend of \pA{}, these methods preferred a large receptive field that guarantees the integration of contextual and local features during prediction and thus eases the generation of the final class map.
\par

\subsection{Performance in \pA{}}

A significant number of submitted methods surpassed the performance of \etal{Ara\'{u}jo}~\cite{Araujo2017}, which reported an overall 4-class accuracy of 77.8\%. Indeed, BACH provided a larger and more representative dataset which, when combined with advances on architectures and transfer-learning techniques, has enabled the development of methods with higher generalization ability. Specifically, these architectures show a high sensitivity for cancer (specially Invasive) cases, which are of great relevance in terms of clinical application (faster automated diagnosis in the cases demanding more urgent attention). Also, as depicted in Table~\ref{tab:pA_sens}~and~\ref{tab:pA_sens_2}, the approaches proposed by the participants outperform simple fine-tuning solutions, indicating that there was a clear effort to improve network performance by changing relevant design and training details. Even though the performance of the top-10 methods is not statistically different (see Fig~\ref{fig:kappa_pA}), a careful design of the experimental setting, including train-validation split, data agumentation, architecture combination and parameter tuning, are essential to increase the performance of deep learning systems.

Despite their high accuracy, the submitted methods still failed on correctly predicting images of the more subtle Benign and \textit{In situ} classes. In fact, Fig. 5b shows that the Benign class is the one that affects the most the performance of the methods, which is to be expected since the presence of normal elements and usual preservation of tissue architecture associated with benign lesions makes this class specially hard to distinguish from normal tissue. Furthermore, the Benign class is the one that presents greater morphological variability and thus discriminant features are more difficult to learn. 

\par
The generalization capacity of the methods is also affected by the image acquisition pipeline. Specifically, during the acquisition of field images, pathologists focus on capturing regions that contain representative features (tissue architecture, cytological features, \textit{etc.}) for the given label. As a consequence, whenever those features are subtle, as it is common on normal tissue, specialists tend to capture non-relevant structures, such as fat cells. Likewise, for \textit{in situ} carcinomas it is common to center the images on mammary ducts, where the cancer is contained. Fig.~\ref{fig:most_erroneous1}~and~Fig.~\ref{fig:most_erroneous1_train} show two images from the test and training sets, respectively. Fig.~\ref{fig:most_erroneous1_train}, labeled as \textit{In situ}, is centered on a duct and surrounded on the left by non-relevant fat tissue. Fig.~\ref{fig:most_erroneous1} has, by coincidence, the same acquisition scheme of Fig.~\ref{fig:most_erroneous1_train} and despite being correctly classified as Benign by 100\% of the experts, 60\% of the top-10 methods classified it as \textit{In situ} and 10\% as Normal. 
Likewise, Fig.~\ref{fig:most_erroneous2} shows a full-consensus Benign test image that was classified by 70\% of the top-10 as Invasive. Once again, this image has a similar overall tissue organization as training cases of other classes, as shown in the invasive tissue depicted in Fig.~\ref{fig:most_erroneous2_train}. The differences, which lie in the cytological features (nuclei size, color and variability), are clear and yet the approaches failed to correctly capture these discriminant characteristics. This suggests that the networks may be partially modeling \textit{how} images were acquired instead of focusing on \textit{what} leads to the classification \cite{Abramoff2016}. 
\par
Finally, Fig.~\ref{fig:reported_performance} shows the difference between the top-10 participants' results reported at the submission time via splitting of the training data (\textit{e.g.} cross-split as train-validation-test) with the achieved performance on the independent test set. The majority of the methods has a  10\% difference over the expected accuracy, showing how important a proper design of a method’s evaluation design is (and how cross-validation scores can be overly optimistic). Namely, this difference may be due to:
\begin{inparaenum}[1)]
\item patient-wise overfit to the training data, \textit{i.e.}, the authors did not had in account the origin of the images when doing the split and, due to the lack of staining normalization, the networks may have memorized specific staining patterns. In fact, Kwok~\cite{Kwok2018} was the only top-10 performer to report a lower expected accuracy. As described in Section~\ref{sec:partAB}, the author performed patient-wise division by clustering images of similar colors, which may contributed to the robustness of the method;
\item over-optimistic train-validation-test split by doing a single split round; and/or
\item excessive hyper parameter-tuning to increase the performance on the split test set, reducing generalization capability.
\end{inparaenum}
With this in mind, future versions of BACH will provide guidelines on data splitting to reduce this discrepancy and improve the overall scientific correctness of the reported results.

\subsubsection{Inter-observer analysis}

The \pA{} BACH dataset was manually annotated by two medical experts and images with dubious diagnosis were discarded. As expected,  the annotator of the dataset tends to be better than his/her peers, which were not capable of correctly classifying, in average, more than 82\% of all images. Consequently, 
in the best scenario the performance of the automatic methods is expected to be equal to that of the observers. Taking into account the average expert accuracy of 85$\pm10\%$, one can see that the performance obtained by the competing solutions (Table~\ref{tab:partA}) is in line with this value, being the highest accuracy of 87$\%$.
 The human-level performance of the algorithms is further corroborated by the partial lack of statistically differences of the methods in comparison to the expert pathologists, as shown in Fig~\ref{fig:kappa_pA}. This is specially true for the scenario of pathology detection, where the participants even outperform one of the experts. However, this difference tends to reduce when considering abnormality detection. In fact, similarly to the automatic methods (recall Fig.~\ref{fig:misclassifications}), the human observers, as shown in Fig.~\ref{fig:conf_A}, have a high agreement level for Invasive cases but tend to disagree on the other classes. Fig~\ref{fig:kappa_pA}~and~\ref{fig:conf_A}, together with Tables~\ref{tab:pA_sens}~and~\ref{tab:pA_sens_2}, suggest that specialists rely not only on objective markers, but also on their experience, intuition and personal assessment of cost of failure to perform the diagnosis, as assessable on the kappa score of the Pathological and Cancer-wise classifications. Also, similarly to the deep learning models, the specialists had more difficulty in distinguishing between the Normal and Benign classes in comparison with the cancerous classes. This further corroborates the hypothesis that the participants tended to fail Benign images due to the previously discussed complexity of this class.
\par
Overall, the results in Table~\ref{tab:pA_sens}~and~~\ref{tab:pA_sens_2}, as well as the comparison of the quadratic Cohen's kappa score (Fig~\ref{fig:kappa_pA}) between the different pathologists \textit{vs} ground-truth and the automatic methods \textit{vs} the ground-truth, show that deep learning models trained on properly annotated data can achieve human performance in complex medical tasks and may in the near future play an important role as second-opinion systems.

\subsection{Performance in \pB{}}

In general, \pB{} is much more challenging than \pA{} due to the large amount of information to process and need to integrate a wide range of scales. The pixel-wise sensitivity and specificity of the methods from \pB{} detailed in Tables~\ref{tab:pB_sens}~and~\ref{tab:pB_sens_2}, shows that the Invasive class tends to be the easiest to detect as the methods achieved an average sensitivity of 0.4. This is to be expected, since Invasive carcinoma is characterizable by an abnormal and non-confined nuclei density and thus methods tend to require less contextual information for the prediction. In fact, this is corroborated by the results in \pA{} that indicate that Invasive is the easiest of the pathological classes (see discussion of Fig.~\ref{fig:misclassifications}). On the other hand, \textit{In situ} has the lowest detection sensitivity (average of 0.06 and maximum value of 0.18) of the pathological classes. Unlike the Invasive carcinoma, classification as \textit{in situ} is reliant on the location of the pathological cells -- which means that without proper global and local context, which is complex to achieve due to the large size of the images, this class becomes non-trivial to classify. On the other hand, the methods of \pA{} did not tend to fail on images from the \textit{in situ} class. This indicates that the microscopy images provide enough local and global context to perform the labeling and thus that human experience had an essential role during the acquisition and annotation of these images.
\par
Fig.~\ref{fig:pb_pred} shows examples of predictions from the top-3 performing participants. In general, one can observe the overestimation of invasive (blue) regions, and more difficulty in predicting the \textit{in situ} (green) and benign (red) ones, which can also be seen in Table~\ref{tab:pB_sens}, where the sensitivity of the solutions for the invasive class is clearly superior to the others. Fig.~\ref{fig:pb_pred} shows this tendency, where two of the top performing teams fail to predict the \textit{in situ} and benign regions and tend estimate them as invasive or as background. 

\subsection{Diversity in the Solutions of the BACH Challenge}

Challenge designs should also promote a higher diversity of methodologies. However, BACH submissions followed the recent computer vision trend with deep learning vastly being the preferred approach. Specifically, pre-trained deep networks on natural images are relatively easy to set up and allow to achieve high performance while significantly reducing the need for field-knowledge, easing the participation on this and other challenges. Although the raw high performance of these methods is of interest, the scientific novelty of the approaches is reduced and usually limited to hyperparameter setup or network ensemble. Also, the black-box behavior of deep learning approaches hinders their application on the medical field, where specialists need to understand the reasoning behind the system's decision. It is the authors' belief that medical imaging challenges should further promote advances on the field by incentivating participants to propose significantly novel solutions that move from \textit{what?} to \textit{why?}. For instance, it would be of interest on future editions to ask participants to produce an automatic explanation of the method's decision. This will require the planning of new ground-truths and metrics that benefit systems that, by providing proper decision reasoning, are more capable of being used in the clinical practice.

\subsubsection{Limitations of the BACH Challenge}
While an effort has been made in creating a relevant, stimulating and fair challenge, capable of advancing the state-of-the-art, the authors are aware of some limitations, namely:
\begin{inparaenum}[1)]
\item
The regional origin and relatively small size of the provided training dataset (specially for deep learning standards) may have limited the generalization capability of the solutions. Likewise, the relatively small test set does not allow to extensively evaluate the behaviour of the algorithms to different tissue structures and staining deviations.  Increasing the diversity of the dataset would allow to draw even more relevant conclusions regarding the performance of image analysis systems for breast cancer diagnosis.  
\item The reference labels for both \pA{} and \pB{} were obtained via manual annotation of two medical experts. Even though images where the observers disagreed were discarded, the labeling process is still reliant on the subjectivity and experience of the annotators (specially on Normal \textit{vs} Benign labeling since no immunohistochemistry analysis is useful), limiting the performance of the submitted methods to that of the human expert. Increasing the number of annotators would allow to further increase the reliability of the dataset.
\item Patient-wise labels were only partially available for the training set. Participants could have used data with known patients for training and the remaining for method validation or use alternative approaches (such as clustering) to estimate the origin of the images. Despite this, the availability of this information for all images would allow a more fair patient-wise split for training and evaluating the algorithms, and could eventually lead to a smaller discrepancy between the training set cross-validated and the test set results. 
\item The pixel-wise annotations of the WSI are not highly detailed and thus the delineated regions may include normal tissue regions besides the class assigned to that region. 
\item Automatic evaluation of the participants' algorithms would have ease the submission procedure, allowing to an almost real-time feedback of the teams' performance. In this scenario, a scheme of multiple submissions could have been implemented, in which teams would be allowed to submit results on the website during the challenge running period, up to a maximum number of submissions. This would probably also boost the number of final submissions out of the challenge registrations.  
\end{inparaenum}

%% file: conclusion.tex
\section{Conclusions}
\label{sec:conclusion}

BACH was organized to promote research on CAD systems for automatic breast cancer histology image analysis. Despite the complexity of the task, the challenge received a large number of high quality solutions that achieve similar performance to human experts. 
Namely, the best performing methods achieved 0.87 accuracy in classifying high resolution microscopy images in Normal, Benign, \textit{In situ} carcinoma and Invasive carcinoma classes and a 0.69 score in labeling entire WSI. 

Proper experiment design seems to be essential to achieve high performance in the breast cancer histology image analysis. Specifically, the conducted study allows to infer that 
\begin{inparaenum}[1)]
\item generically, using the latest CNN designs allows to positively impact the system's performance given that fine-tuning is properly performed;
\item CNNs seem to be robust to small color variations of H\&E images and thus color normalization was not essential to attain high accuracies;
\item proper training splitting is essential to infer the generalization capability of the model, since CNNs may overfit to patient/acquisition details, and
\item using large context images as the network input allows for overall high performance even if the input image (and thus the overall quality of the information) has to be downsampled.
\end{inparaenum}
On the other hand, current deep learning solutions still have issues dealing with large, high resolution images and further investment on development of methods for WSI analysis should be done. 
\par

It is the organizaners' hope that the comprehensive analysis herein presented will motivate more challenges on medical imaging and specially pave the way for the development of new breast cancer CAD methods that contribute to the early detection of this pathology, with clear benefits for our societies.